%% file: main.tex
\documentclass[sigconf]{acmart} %

\usepackage{float}
\usepackage{enumitem}
\usepackage[ruled, linesnumbered, vlined]{algorithm2e}

\usepackage{graphicx} 
\usepackage{epstopdf}
\usepackage{caption}
\usepackage{amsmath,bm}
\usepackage{ulem}
\usepackage{multirow}
\usepackage{bbm}
\usepackage{tabularx}
\usepackage{multirow}
\usepackage{array}
\usepackage{tcolorbox}
\usepackage{xcolor}
\usepackage{soul} 
\usepackage{float}

\DeclareMathOperator*{\argmax}{arg\,max}

\newcommand{\Comment}{}

\AtBeginDocument{%
  \providecommand\BibTeX{{%
    \normalfont B\kern-0.5em{\scshape i\kern-0.25em b}\kern-0.8em\TeX}}}

\copyrightyear{2025}
\acmYear{2025}
\setcopyright{cc}
\setcctype{by}
\acmConference[SIGIR '25]{Proceedings of the 48th International ACM SIGIR Conference on Research and Development in Information Retrieval}{July 13--18, 2025}{Padua, Italy}
\acmBooktitle{Proceedings of the 48th International ACM SIGIR Conference on Research and Development in Information Retrieval (SIGIR '25), July 13--18, 2025, Padua, Italy}\acmDOI{10.1145/3726302.3730009}
\acmISBN{979-8-4007-1592-1/2025/07}

\newtcolorbox{highlighted}[1][]{
    colback=#1!30,      
    colframe=#1!50,     
    boxrule=0mm,        
    sharp corners,      
    boxsep=0pt,         
    left=1pt, right=1pt, top=1pt, bottom=1pt, 
    valign=center,      
    box align=base,     
    boxrule=0pt,        
    halign=flush left,  
    fit to height=3.5em, 
    fit to text,        
}

\begin{document}

\title{InstructRAG: Leveraging Retrieval-Augmented Generation on Instruction Graphs for LLM-Based Task Planning}

\author{Zheng Wang}
\authornote{Equal Contribution.}
\affiliation{%
  \institution{Huawei Singapore Research Center}
  \country{Singapore}}
\email{wangzheng155@huawei.com}

\author{Shu Xian Teo}
\authornotemark[1]
\affiliation{%
  \institution{Huawei Singapore Research Center}
  \country{Singapore}}
\email{teo.shu.xian@huawei.com}

\author{Jun Jie	Chew}
\affiliation{%
  \institution{Huawei Singapore Research Center}
  \country{Singapore}}
\email{chew.jun.jie@huawei.com}

\author{Wei Shi}
\affiliation{%
  \institution{Huawei Singapore Research Center}
  \country{Singapore}}
\email{w.shi@huawei.com}

\renewcommand{\shortauthors}{Zheng Wang, Shu Xian Teo, Jun Jie Chew, Wei Shi}

\input{abstract}

\ccsdesc[500]{Information systems~Information retrieval}

\keywords{large language model, retrieval-augmented generation, agent planning}
\maketitle

\input{introduction}
\input{related}
\input{problem}
\input{method}

\input{setup}
\input{experiments}

\input{conclusion}

\bibliographystyle{ACM-Reference-Format}
\bibliography{ref}

\clearpage
\input{appendix2}

\end{document}

%% file: abstract.tex
\begin{abstract}

Recent advancements in large language models (LLMs) have enabled their use as agents for planning complex tasks. Existing methods typically rely on a thought-action-observation (TAO) process to enhance LLM performance, but these approaches are often constrained by the LLMs' limited knowledge of complex tasks. Retrieval-augmented generation (RAG) offers new opportunities by leveraging external databases to ground generation in retrieved information.
In this paper, we identify two key challenges (enlargability and transferability) in applying RAG to task planning. We propose \texttt{InstructRAG}, a novel solution within a multi-agent meta-reinforcement learning framework, to address these challenges. \texttt{InstructRAG} includes a graph to organize past instruction paths (sequences of correct actions), an RL-Agent with \underline{R}einforcement \underline{L}earning to expand graph coverage for enlargability, and an ML-Agent with \underline{M}eta-\underline{L}earning to improve task generalization for transferability. The two agents are trained end-to-end to optimize overall planning performance. Our experiments on four widely used task planning datasets demonstrate that \texttt{InstructRAG} significantly enhances performance and adapts efficiently to new tasks, achieving up to a {\Comment 19.2\%} improvement over the best existing approach.

\end{abstract}

%% file: introduction.tex
\section{INTRODUCTION}
\label{sec:introduction}


With the significant advancement of large language models (LLMs), a recent trend has emerged in employing LLMs as intelligent agents to tackle diverse real-world planning tasks. These tasks include multi-hop reasoning~\cite{yang2018hotpotqa}, embodied tasks~\cite{shridhar2020alfworld, wang2024divscene, shridhar2020alfred}, web shopping~\cite{yao2022webshop}, and scientific reasoning~\cite{wang2022scienceworld}, etc. Many recent solutions to the planning problem, such as ReAct~\cite{yao2022react}, KnowAgent~\cite{zhu2024knowagent}, WKM~\cite{qiao2024agent}, Reflexion~\cite{shinn2024reflexion}, FireAct~\cite{chen2023fireact}, NAT~\cite{wang2024learning}, and ETO~\cite{song2024trial}, follow a thought-action-observation (TAO) process. In the \underline{thought} phase, the LLM leverages its reasoning ability to create a plan by breaking down a task into a series of subtasks. In the \underline{action} phase, the LLM determines the specific actions required, such as selecting which tool to use. In the \underline{observation} phase, it captures the results of executing the action and provides feedback from the external environment to the LLM, facilitating the planning of subsequent TAO steps. Within this process, the thought and action are generated by the LLM, while the observation is implemented by the environment. Existing solutions adopt diverse strategies, including prompting~\cite{yao2022react, shinn2024reflexion} or fine-tuning~\cite{zhu2024knowagent, qiao2024agent, chen2023fireact, wang2024learning, song2024trial}, to improve LLM-generated thoughts and actions for more effective planning. In particular, KnowAgent~\cite{zhu2024knowagent} integrates pre-defined rules into prompts to ensure that generated thoughts exhibit logical action transitions. For example, it prevents looking up an entity without first performing a search operation on the topic, as seen in HotPotQA~\cite{yang2018hotpotqa}. Reflexion~\cite{shinn2024reflexion} incorporates self-reflection summaries into the TAO process to guide subsequent trials. WKM~\cite{qiao2024agent} trains a world knowledge model to generate thoughts based on knowledge acquired from human task-solving experiences.

While these existing methods aim to enhance LLM planning, they are often constrained by the inherent limitations of the LLMs themselves, such as their limited knowledge on complex tasks. The rapid development of retrieval-augmented generation (RAG) provides new opportunities to address these limitations by leveraging external databases. By anchoring LLM generation in retrieved information, RAG improves performance through the integration of relevant data during the planning process. In this context, we recognize that task-specific nature of information retrieval plays a crucial role for effective planning generation.
For example, consider the question ``Were Scott Derrickson and Ed Wood of the same nationality?'' from HotPotQA, as shown in Figure~\ref{fig:overall}(a). A potential retrieved plan for this question involves a sequence of actions (referred to as instruction paths in this paper): first, use Google Search to find information about Scott Derrickson (denoted by \texttt{Search[Scott Derrickson]}), then look up a sentence containing the keyword ``nationality'' (\texttt{Lookup[nationality]}), followed by \texttt{Search[Ed Wood]} and \texttt{Lookup[nationality]}. These instructions are specific to the question at hand and may vary depending on the topics or entities involved. A similar phenomenon can also be observed in ALFWorld embodied tasks, where instructions might include \texttt{Goto[shelf 6]}, then \texttt{Take[vase 2 from shelf 6]}.

In this paper, we discuss the task-specific nature in two aspects, with the goal of bridging the gap between task-specific questions and instructions derived from past experiences stored in a database through RAG. 
\underline{(1) Enlargeability}: This refers to a task where the question falls \emph{within} the scope of those covered by the external database. Specifically, we pre-store successful instruction paths in the database, with each path tailored to a specific task. To address questions related to these tasks, we explore a paradigm for combining instructions to expand the database's coverage. 
As illustrated in Figure~\ref{fig:overall}(a), there are two successful instruction paths, $P_1^1$ and $P_2^1$, for solving questions $Q_1^1$ and $Q_2^1$, respectively. These paths consist of five instructions: searching for (a) Scott Derrickson, (b) Ed Wood, and (c) Christopher Nolan, and looking up entities related to (d) nationality and (e) birthplace. By combining these instructions in sequences such as $(a) \rightarrow (e) \rightarrow (c) \rightarrow (e)$, we can generate a new instruction path for solving a novel question (i.e., $\hat{Q}$) that was not covered by the original paths (but shares the same task of querying a location). The enlargeability is proposed to enhance the database's ability to address a wider range of questions. 
\underline{(2) Transferability}: This refers to a task where the question is \emph{outside} the scope of tasks covered by the external database. We note that LLM-based task planning is provided as a capability to support a wide range of tasks in practice. Transferability is essential for bridging the gaps between different tasks (i.e., those in the pre-built database and the questions at hand) within the RAG system. To achieve this, it is necessary to expand the database to incorporate new instructions required for different tasks, such as updating it with instructions relevant to the new tasks based on a development set. Additionally, certain trainable modules associated with the RAG can be rapidly adapted to accommodate the new tasks. 

\noindent\textbf{New Solution.} Although recent research efforts~\cite{lee2024planrag,kagaya2024rap,shi2024generate} have attempted to apply RAG techniques to task planning, these methods often fall short in several aspects: i)~\cite{lee2024planrag} is primarily tailored for specific domains, such as decision-making in video games, making it challenging to generalize their designs to broader planning tasks as studied in this paper. ii)~\cite{shi2024generate} focuses on multi-hop reasoning via search engines (e.g., using Google Search to access Wikipedia knowledge), but their effectiveness in tasks where the search engines are inapplicable (e.g., embodied tasks or web shopping) remains unexplored. iii)~\cite{kagaya2024rap} simply relies on storing past experiences and retrieving similar ones using AKNN, without identifying key aspects such as enlargability and transferability. This gap results in suboptimal performance, as evidenced by our experiments.

To this end, we propose \texttt{InstructRAG}, a new solution based on a multi-agent meta-reinforcement learning framework.
\underline{For (1)}, we design an instruction graph to instantiate the database. In this graph, nodes and edges represent two sets: nodes contain similar instructions, while edges represent corresponding tasks, all derived from successful instruction paths in past experiences. The rationale behind this approach is two-fold: 1) The graph provides a natural structure for organizing paths and facilitates the integration of new paths by clustering similar instructions related to various tasks. 2) Each node acts as a junction that enables the creation of new paths by combining stored instructions within it, and each edge records the tasks (with associated questions) along the path. This organization allows us to structure past experiences effectively within the database. Further, we design an RL-Agent that utilizes \underline{R}einforcement \underline{L}earning to identify candidate paths on the graph, with the goal of optimizing the database’s coverage to enhance its enlargeability.
\underline{For (2)}, we explore a meta-learning approach into the RAG pipeline. Specifically, we introduce an additional agent, referred to as the ML-Agent, which \underline{M}eta-\underline{L}earns to select a path from the candidate paths provided by the RL-Agent. This selected path is then used as an in-context learning exemplar within the prompt, aiming to enhance the LLM's generalization to new tasks by updating it with only a few QA pairs during the meta-update phase. Here, the two agents collaborate within the TAO process~\cite{yao2022react} to facilitate task planning via grounding the generation of thoughts and actions by LLMs. We note that the RL-Agent generates candidate paths for the ML-Agent to select, and the ML-Agent then assesses the end-to-end effectiveness of the selected path, to incorporate this feedback as the reward for the RL-Agent. This interaction creates a positive loop, leading to improved planning performance.

To summarize, we make the following contributions. 
\begin{itemize}[leftmargin=3mm]
    \item We conduct a systematic study of leveraging RAG for LLM-based task planning, and identify two key properties (i.e., enlargability and transferability) that a potential technique should possess. To our best knowledge, this is the first attempt of its kind.
    \item We propose a new solution called \texttt{InstructRAG}, which includes three key components: an instruction graph, an RL-Agent, and an ML-Agent. These components are integrated into a multi-agent meta-reinforcement learning framework that explicitly trains to optimize end-to-end task planning performance.
    \item We conduct extensive experiments on four widely used task planning datasets: HotpotQA~\cite{yang2018hotpotqa}, ALFWorld~\cite{shridhar2020alfworld}, Webshop~\cite{yao2022webshop}, and ScienceWorld~\cite{wang2022scienceworld}, across three typical LLMs. Our \texttt{InstructRAG} can be integrated with both trainable LLMs (e.g., GLM-4~\cite{glm2024chatglm}) for fine-tuning and frozen LLMs (e.g., GPT-4o mini~\cite{achiam2023gpt} and DeepSeek-V2~\cite{deepseekv2}). The results demonstrate that \texttt{InstructRAG} improves performance by approximately {\Comment 19.2\%}, {\Comment 9.3\%}, {\Comment 6.1\%}, and 10.2\% over the best baseline method on the four datasets, respectively. In addition, \texttt{InstructRAG} adapts rapidly to new tasks, achieving effective performance with few-shot learning.
\end{itemize}

%% file: related.tex
\section{RELATED WORK}
\label{sec:related}

\noindent\textbf{LLM-based Agent Planning.} To solve complex tasks, humans typically decompose them into smaller sub-tasks and then evaluate the plan's effectiveness. Similarly, LLM-based agents follow this routine, and we categorize existing techniques based on whether the agent receives feedback during the planning process. A detailed survey of LLM-based agent planning can be found in~\cite{wang2024survey,xi2023rise}. \emph{In planning without feedback}, agents do not receive feedback that influences their future actions. The main techniques for this category include (1) single-path reasoning~\cite{wei2022chain,raman2022planning}, (2) multi-path reasoning~\cite{wang2022self,yao2024tree,besta2024graph}, and (3) using an external planner~\cite{liu2023llm+,dagan2023dynamic}. Specifically, for (1), CoT~\cite{wei2022chain} illustrates the reasoning steps for LLMs to tackle complex tasks using prompts, thereby guiding LLMs to plan and execute actions step-by-step. For (2), CoT-SC~\cite{wang2022self} explores diverse reasoning paths to solve complex tasks. Initially, it utilizes CoT to generate multiple reasoning paths and their respective answers. Subsequently, it selects the answer with the highest frequency as the final output. For (3), external planners are designed to generate plans for specific domains. For example, LLM+P~\cite{liu2023llm+} focuses on robot planning tasks by defining them using formal Planning Domain Definition Languages (PDDL). It utilizes an external planner, such as the Fast Downward planner~\cite{helmert2006fast}, which employs heuristic search to handle PDDL formulations. The results generated by the planner are then translated back into natural language by LLMs.

\emph{In planning with feedback}, effectiveness is generally improved by receiving feedback after actions are taken, which supports long-horizon planning. This feedback can come from (1) environments~\cite{yao2022react,chen2023fireact}, (2) humans~\cite{qiao2024agent,zhu2024knowagent}, and (3) models~\cite{shinn2024reflexion,madaan2024self}. For (1), ReAct~\cite{yao2022react} proposes the TAO process, where a language model generates the thought for planning, the action involves interacting with the environment, and the observation consists of external feedback (such as search engine results) based on the action. FireAct~\cite{chen2023fireact} generates the TAO using various methods, which are then converted into the ReAct format to fine-tune a small language model. For (2), KnowAgent~\cite{zhu2024knowagent} integrates action knowledge, which includes rules determining action transitions, into prompts to enhance the planning capabilities of LLMs. This knowledge is derived from both human input and GPT-4~\cite{achiam2023gpt}. Further, WKM~\cite{qiao2024agent} is introduced to facilitate agent planning using a world knowledge model. This model is trained by comparing selected trajectories (annotated by humans) with rejected trajectories (explored by an experienced agent). For (3), Reflexion~\cite{shinn2024reflexion} employs verbal feedback to enhance the agent's planning based on previous failures. It transforms binary or scalar feedback from self-evaluation into a textual summary, which is then added as additional context for the agent in subsequent planning. 
In this paper, we explore a new RAG-based approach to task planning, emphasizing two key properties: enlargability and transferability in technique development.

\smallskip\noindent\textbf{Retrieval-Augmented Generation.} RAG enhances LLM generation by querying an external database to obtain relevant information, which grounds the subsequent text generation.
Recent studies utilize RAG for task planning~\cite{wang2024rat,kagaya2024rap,lee2024planrag,shi2024generate}. Specifically, RAT~\cite{wang2024rat} enhances CoT by iteratively revising each thought step with retrieved information relevant to the task query, thereby improving LLMs' ability to reason over long-horizon generation tasks. 
RAP~\cite{kagaya2024rap} stores past experiences, including context and action-observation trajectories, and retrieves them based on their similarity to the current situation. The goal is to facilitate deriving appropriate actions by leveraging memory examples from similar tasks.
PlanRAG~\cite{lee2024planrag} is designed for decision QA tasks, following a plan-then-retrieval approach. The LLM first generates a plan to guide the analysis, then retrieves information from an external database by formulating queries. It also continuously evaluates the need for re-planning during the process.
%
GenGround~\cite{shi2024generate} explores a generate-then-ground approach for multi-hop reasoning tasks. It breaks down a complex question into sub-questions, generates an immediate answer for each, then revises it with retrieved information. This revised answer informs the next sub-question, iterating until the final answer is achieved. In this paper, we propose \texttt{InstructRAG} within a multi-agent meta-reinforcement learning framework to systematically address the gap in leveraging RAG for task-specific questions and stored past experiences.

\smallskip\noindent\textbf{Meta-learning for Improving LLMs via In-context Learning (ICL).} To enhance the transferability of LLMs to unseen tasks, meta-learning approaches~\cite{min2021metaicl, chen2022meta, sinha2024maml, deb2022boosting} have been developed. These approaches fine-tune pre-trained LLMs using a diverse set of tasks, formatted as ICL instances by pre-appending task-specific exemplars to the prompts during training. These methods follow Model-Agnostic Meta-Learning (MAML) principles~\cite{finn2017model}. Specifically, MAML-en-LLM~\cite{sinha2024maml} explores a wide parameter space to learn truly generalizable parameters that perform well on disjoint tasks and adapt effectively to unseen tasks. MTIL~\cite{deb2022boosting} investigates the application of meta-learning to multi-task instructional learning~\cite{wang2022benchmarking}, aiming to enhance generalization to unseen tasks in a zero-shot setting. MetaICL~\cite{min2021metaicl} adapts a LLM to perform in-context learning across a broad range of training tasks. It aims to improve the model’s ability to learn new tasks in context during testing, by conditioning on a few training examples without requiring parameter updates or task-specific templates. In this paper, we propose a novel meta-reinforcement learning framework to improve transferability, with two cooperative agents tailored for planning tasks. This approach is distinctly different from existing methodologies in the field.

%% file: problem.tex
\section{PROBLEM STATEMENT}
\label{sec:problem}

We explore the problem of LLM-based task planning through RAG, grounded in an external database (i.e., instruction graph). In this context, we identify two practical properties that should be met:

\begin{itemize}[leftmargin=3mm]
    \item[-] Enlargeability: It should expand the scope of the instruction graph by traversing existing instructions (nodes) on the graph and combining them into new sequences of instructions (paths). This will help the LLM in completing tasks that do not have pre-defined paths during the graph's construction.
    \item[-] Transferability: Task planning as a capability in practice involves developing techniques that achieve rapid adaptation to new tasks. For example, the trained model should be able to quickly learn a new task from a small amount of new data. 
\end{itemize}

%% file: method.tex
\section{METHODOLOGY}
\label{sec:method}

\begin{figure*}
  \centering
  \includegraphics[width=\linewidth]{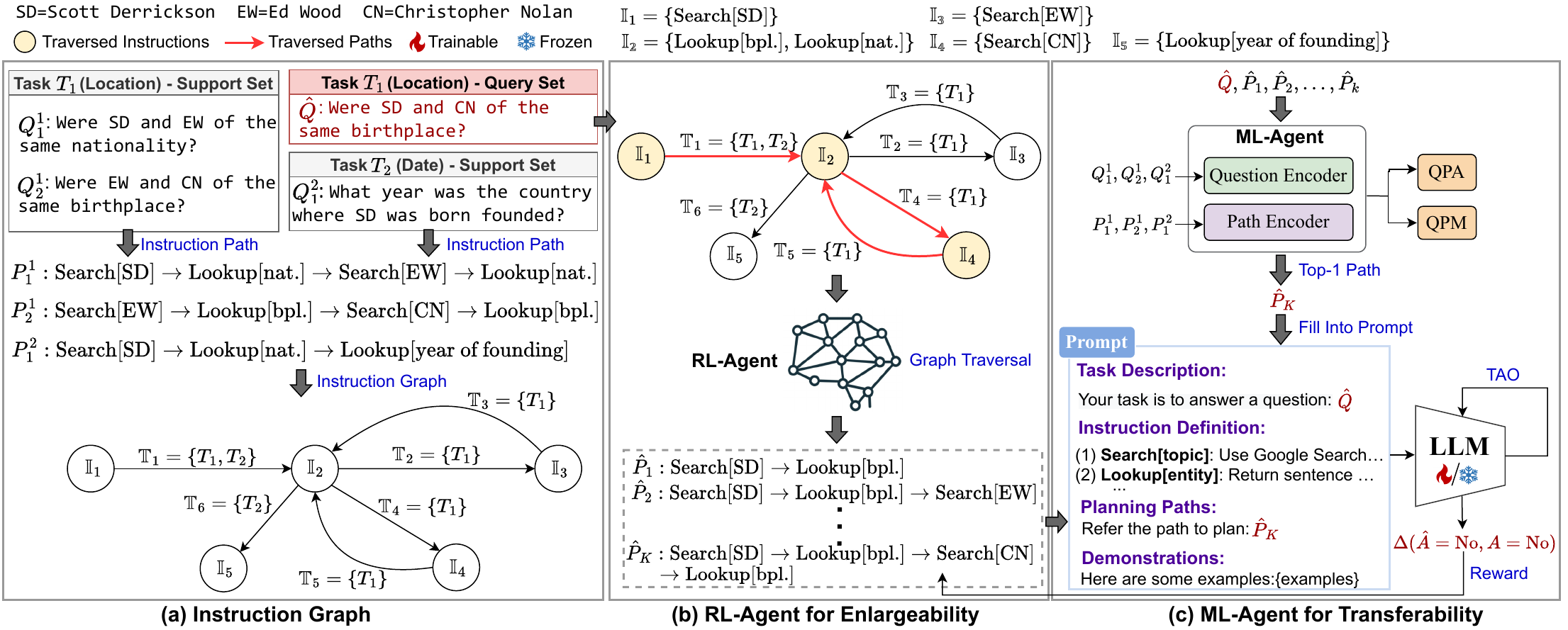}
  \vspace*{-4mm}
  \caption{Architecture of the proposed \texttt{InstructRAG} with multi-agent meta-reinforcement learning, illustrated on HotpotQA.}\label{fig:overall}
  \vspace*{-4mm}
\end{figure*}

\subsection{Overview of \texttt{InstructRAG}}
\label{sec:overview}
The proposed \texttt{InstructRAG} tackles the challenges of LLM-based task planning by focusing on two key properties: enlargeability and transferability. It comprises several components: instruction graph construction (Section~\ref{sec:graph}), RL-Agent (Section~\ref{sec:rlagent}), and ML-Agent (Section~\ref{sec:mlagent}). These components are integrated into a multi-agent meta-reinforcement learning framework, which is detailed in terms of three stages: training, few-shot learning, and testing (Section~\ref{sec:instructrag}). 

\smallskip\noindent\textbf{Training}. We illustrate the overall framework in Figure~\ref{fig:overall}. Specifically, the training tasks (seen tasks) are divided into a support set and a query set. \underline{For support set}, it is used to construct the instruction graph by extracting instruction paths from questions and forming the graph based on these paths. Additionally, the paths and corresponding questions help warm-start the RL-Agent, and pre-train the ML-Agent with two pre-training tasks: Question Path Alignment (QPA) and
Question Path Matching (QPM). \underline{For query set}, it is used to query the graph and trains the RL-Agent and the ML-Agent within a multi-agent framework. The RL-Agent finds candidate paths through graph traversal, which is modeled as a Markov Decision Process (MDP) and optimized using reinforcement learning. The RL-Agent is trained to handle questions not seen during the graph construction, enlarging the capability of the instruction graph by combining instructions into the paths to address these questions. The ML-Agent then selects the most relevant path among the candidate paths based on their representations, which is used to form the prompt for a LLM to predict the final answer, following the TAO process. We note that the transferability is considered through an in-context learning manner, where the LLM learns a new task by conditioning on the task-specific path in the prompt. The ML-Agent optimizes this process, either with a trainable or frozen LLM, via meta-learning.

\smallskip\noindent\textbf{Few-shot Learning and Testing}. Once the parameters of the RL-Agent and ML-Agent are meta-trained, we rapidly adapt the model parameters using few-shot examples from the support set on testing tasks (unseen tasks). The testing is then conducted based on the query set of these testing tasks.

\subsection{Instruction Graph}
\label{sec:graph}

\smallskip
\noindent\textbf{Instruction.} An instruction $I$ represents a specific action performed by LLMs, e.g., \texttt{Search[Scott Derrickson]} is an instruction, meaning to use Google Search to find relevant information about Scott Derrickson as shown in Figure~\ref{fig:overall}(a). 

\smallskip
\noindent\textbf{Instruction Path.} An instruction path $P_i^j = \langle I_1, I_2, \ldots, I_{|P|} \rangle$ is represented as a sequence of instructions that LLMs follow step-by-step to perform actions and complete the $i$-th question of the $j$-th task, e.g., $P_1^2$: \texttt{Search[Scott Derrickson]} $\rightarrow$ \texttt{Lookup[nationality]} $\rightarrow$ \texttt{Lookup[year of founding]} denotes an instruction path to address the question $Q_1^2$ of the task $T_2$ as shown in Figure~\ref{fig:overall}(a). 

\smallskip
\noindent\textbf{Instruction Graph.} An instruction graph $G(\mathbb{V},\mathbb{E})$ is represented as a directed graph that organizes instruction paths of questions belonging to various tasks, where $\mathbb{V}$ and $\mathbb{E}$ represent the nodes and edges of the graph, respectively. Each node $\mathbb{I} \in \mathbb{V}$ denotes an instruction set, i.e., $\mathbb{I} = \{I_1, I_2, \ldots, I_{|\mathbb{I}|}\}$, clustering similar instructions. Each edge $\mathbb{T} \in \mathbb{E}$ denotes a task set, i.e., $\mathbb{T} = \{T_1, T_2, \ldots, T_{|\mathbb{T}|}\}$, recording the tasks with associated questions involved on the path.

\smallskip
\noindent\textbf{The Graph Construction and Insights.} We present the graph construction in two steps, illustrated with a running example in Figure~\ref{fig:overall}(a). The detailed process is outlined in Algorithm~\ref{alg:ig}.

\underline{Step-1 (Generating Instruction Paths)}: We divide the dataset into two parts: the support set and the query set, following the meta-learning setup. The support set is used for graph construction, while the query set is used to query the graph to train enlargeability and transferability, to be discussed in Section~\ref{sec:rlagent} and Section~\ref{sec:mlagent}, respectively. For each question in the support set, we generate its instruction path using existing task planning techniques~\cite{zhu2024knowagent,yao2022react,shinn2024reflexion}. We select the path that correctly plans the question for construction, ensuring the planning is grounded in the prepared database, aligning with the goal of RAG.
    
\underline{Step-2 (Inserting Instructions with a Threshold $\delta$)}: Then, we iteratively insert each instruction in the generated paths, i.e., $P_1^1, P_2^1, P_1^2$, into the graph $G$. The first two instructions in $P_1^1$ are initialized to create two node sets, i.e., $\mathbb{I}_1 \leftarrow$ \texttt{Search[Scott Derrickson]} and $\mathbb{I}_2 \leftarrow$ \texttt{Lookup[nationality]}, which correspond to an edge set $\mathbb{T}_1$ recording the involved task $\{T_1\}$. Here, we note that adjacent instructions are not inserted into the same node, as this would break the transition between them.
To insert the next instruction \texttt{Search[Ed Wood]}, we perform an AKNN search~\cite{malkov2018efficient} on all instructions excluding the instructions in its adjacent node (i.e., \texttt{Lookup[nationality]} $\in \mathbb{I}_2$), identifying the most similar instruction \texttt{Search[Scott Derrickson]}, which is associated with a similarity value (e.g., cosine similarity) $\psi$ in the node set $\mathbb{I}_1$. Then, we define a threshold $\delta$ to control the insertion. If $\psi < \delta$, a new node set $\mathbb{I}_3$ is created and the instruction is inserted into this new node, i.e., $\mathbb{I}_3 \leftarrow$ \texttt{Search[Ed Wood]}; otherwise, the instruction is inserted into the identified node $\mathbb{I}_1$. The process continues until all instructions are inserted. 
Additionally, we note that when the instruction \texttt{Lookup[nationality]} of $P_1^2$ is inserted into $\mathbb{I}_2$ (with a cosine similarity of 1.0), the task $T_2$ is also added to the edge set $\mathbb{T}_1$, resulting in $\{T_1, T_2\}$.


We present two key \underline{insights} into graph construction: (1) Graphs naturally organize instruction paths, where nodes and edges are represented as sets to enable flexible integration of similar instructions across tasks. (2) The threshold controls instruction similarity, forming junction nodes that create new paths beyond those in the original data. For instance, merging \texttt{Lookup[nationality]} and \texttt{Lookup[birthplace]} into $\mathbb{I}_2$ enables novel paths like $\mathbb{I}_1 \rightarrow \mathbb{I}_2 \rightarrow \mathbb{I}_4 \rightarrow \mathbb{I}_2$, thus improving graph expandability to cover more questions (e.g., $\hat{Q}$ in Figure~\ref{fig:overall}(a)).


\SetKwInOut{KwIn}{Require}
\begin{algorithm}
    \small
    \caption{The Instruction Graph Construction}
    \label{alg:ig}
	\KwIn{
        a support set $\mathbb{S}$; a threshold $\delta$}
        $IC \leftarrow 3$, $TC \leftarrow 2$ // two counters for node and edge sets \\ 
        \For {each $T_j \in \mathbb{S} (1 \le j \le |\mathbb{S}|$)} { 
       
        \For{each $Q_i^j \in T_j (1 \le i \le |T_j|$)} {
            obtain a correct $P_i^j = \langle I_1, I_2, \ldots, I_{|P_i^j|} \rangle$ for $Q_i^j$\\
            $\mathbb{I}' \leftarrow \emptyset$ // record the last node set\\
            \For{$k=1,2,...,|P_i^j|$}{
            \uIf{$i=1$ and $j=1$ and $k<3$}{
            $\mathbb{I}_1.\text{add}(I_1)$, $\mathbb{I}_2.\text{add}(I_2)$,
            $\mathbb{T}_1 \leftarrow\text{Edge}(\mathbb{I}_1, \mathbb{I}_2)$\\
            $\mathbb{T}_1.\text{add}(T_1)$, $G.\text{addEdge}(\mathbb{T}_1)$, $\mathbb{I}' \leftarrow \mathbb{I}_2$\\
            \textbf{continue}
            }
            recall $\mathbb{I}_s$ and $\psi$ for $I_k$ with AKNN on $G.\mathbb{V} - \mathbb{I}'$\\
            \uIf{$\psi < \delta$}{
            $\mathbb{I}_{IC}.\text{add}(I_k)$, $\mathbb{T}_{TC} \leftarrow\text{Edge}(\mathbb{I}',\mathbb{I}_{IC})$\\
            $\mathbb{T}_{TC}.\text{add}(T_j)$ , $G.\text{addEdge}(\mathbb{T}_{TC})$\\
            $\mathbb{I}' \leftarrow \mathbb{I}_{IC}$, $IC \leftarrow IC + 1$, $TC \leftarrow TC + 1$\\
            }
            \uElse{
                $\mathbb{I}_s.\text{add}(I_k)$\\
                \uIf{$\text{Edge}(\mathbb{I}', \mathbb{I}_s) \in G$}{
                obtain the edge of $(\mathbb{I}', \mathbb{I}_s)$ denoted by $\mathbb{T}'$\\
                $\mathbb{T}'.\text{add}(T_j)$\\
                }
                \uElse{
                $\mathbb{T}_{TC} \leftarrow \text{Edge}(\mathbb{I}', \mathbb{I}_s)$, $\mathbb{T}_{TC}.\text{add}(T_j)$\\ 
                $G.\text{addEdge}(\mathbb{T}_{TC})$, $TC \leftarrow TC + 1$\\
                }
                $\mathbb{I}' \leftarrow \mathbb{I}_s$\\
            }
            }
        }
    }
\textbf{Return} the instruction graph $G$
\end{algorithm}

\subsection{RL-Agent: Retrieving Instruction Paths on Instruction Graph}
\label{sec:rlagent}
Given an instruction graph $G$, we explore its enlargeability through graph traversal to retrieve various instruction paths that solve questions denoted by $\hat{Q}$ not present during construction (i.e., questions in the query set). To achieve this, we train an agent for traversal, which examines each path in the graph, e.g., via depth-first search (DFS). For each node, the agent decides whether to include or exclude the node (i.e., actions) in the path based on the instructions contained in the node and the tasks connected by its edges (i.e., states). A high-quality retrieved path benefits subsequent planning, reflected by an end-to-end metric such as F1 scores on HotPotQA~\cite{yang2018hotpotqa} (i.e., rewards), which can then inform instruction selection. This process forms a Markov Decision Process (MDP), and we employ Reinforcement Learning (RL) to optimize it. 

\smallskip
\noindent\textbf{Constructing Decision Environment.} The instruction graph $G$ typically contains numerous instruction paths, formed by combining different instructions at each node. To manage this, we limit the RL-Agent's retrieval to $K$ relevant instruction paths, denoted as $\hat{P}_1, \hat{P}_2, \ldots, \hat{P}_K$, which are then utilized for planning in the next phase by the ML-Agent (to be introduced in Section~\ref{sec:mlagent}), where the $K$ is a hyperparameter that can be tuned for optimal performance. We first perform an AKNN search on all instructions for a query $\hat{Q}$. The agent's traversal starts from the most similar instructions (corresponding to the nodes) using DFS. Once the agent excludes a node and backtracks to another branch, an instruction path is formed. This process continues until $K$ paths are retrieved.

\smallskip
\noindent\textbf{States.} 
Suppose we have an input question $\hat{Q}$ and visit a node $\mathbb{I}$ (a set of instructions), along with its in-degree edge $\mathbb{T}$ (a set of tasks). We define the state $\mathbf{s}$ using three cosine similarities $CS(\cdot,\cdot)$, that is

\begin{equation}
\begin{aligned}
\label{eq:state}
& \mathbf{s} = \{\max\limits_{I_i \in \mathbb{I}}CS(\mathbf{v}_{\hat{Q}}, \mathbf{v}_{I_i}), 
               \max\limits_{T_j \in \mathbb{T}}CS(\mathbf{v}_{\hat{Q}}, \mathbf{v}_{T_j}), \max\limits_{Q_k^c \in T_c}CS(\mathbf{v}_{\hat{Q}}, \mathbf{v}_{Q_k^c})\},\\
& \mathbf{v}_{T_j} = \frac{1}{|T_j|} \sum_{k=1}^{|T_j|} \mathbf{v}_{Q_k^j} \; \text{and} \;  c = \argmax\limits_{T_c \in \mathbb{T}}CS(\mathbf{v}_{\hat{Q}}, \mathbf{v}_{T_c}),
\end{aligned}
\end{equation}
where $\mathbf{v}_{\cdot}$ denotes an embedding vector. We construct the state by (1) examining the most similar instruction in the node, (2) identifying the most similar task, denoted by $T_c$, in the edge (whose embedding is calculated as the average of the question embeddings belonging to the task), and (3) finding the most similar question within $T_c$.

\smallskip
\noindent\textbf{Actions.} Let $a$ denote an action, which has two choices during the graph traversal: including the visited node by selecting the most similar instruction into the path $\hat{P}_i$ ($1 \le i \le K$) and searching its connected nodes, or excluding the node and backtracking the search from another branch, then an instruction path is formed. The action $a$ is formally defined as:

\begin{align}
\label{eq:action}
a = 1 \text{ (including)}\ \text{or} \ 0 \text{ (excluding)}.
\end{align}    
Considering the consequence of performing an action, it transitions the environment to the next state $\mathbf{s}'$, affecting which node or edge is used for constructing the state. Notably, some predefined rules may be further incorporated to constrain the action space (e.g., a rule of avoiding \texttt{Lookup} information without first performing \texttt{Search} on HotpotQA~\cite{zhu2024knowagent}), which benefits more accurate path selection.

\smallskip
\noindent\textbf{Rewards.} Let $r$ denote a reward, which corresponds to the end-to-end feedback of an instruction path that contributes to the generated answer $\hat{A}$ by a LLM for $\hat{Q}$. Specifically, when an instruction path from the $K$ paths is selected and written into the prompt by the ML-Agent, the LLM generates an answer $\hat{A}$. This answer can be evaluated using a specific metric $\Delta(\cdot,\cdot)$ (e.g., F1 score), defined as:

\begin{align}
\label{eq:reward_r}
r = \Delta(\hat{A}, A),
\end{align}
where $A$ denotes the ground truth answer. The rationale for designing the reward is to enable joint optimization for the two agents in a multi-agent setup, where the RL-Agent provides paths for the ML-Agent to write into the prompt, and the feedback from the prompt affects the path retrieval by the RL-Agent. Therefore, the two agents can be jointly optimized to improve the overall performance.

\smallskip
\noindent\textbf{Policy Learning.} We involve two phases for training the MDP policy: warm-start (WS) and policy gradient (PG). \underline{In WS}, the goal is to equip the agent with the basic ability to include or exclude instructions. To achieve this, we randomly sample questions from the support set. For each question, we randomly sample nodes on $G$ and construct its state using Equation~\ref{eq:state}. If the node is on the instruction path for the question, the state is associated with an action labeled 1; otherwise, it is labeled 0. We collect these state-action pairs and train the RL-Agent using binary cross-entropy:
\begin{align}
\label{eq:ws}
\mathcal{L}_\text{WS} = - y * \log(P) + (y - 1) * \log(1 - P),
\end{align}
where $y$ denotes the label, and $P$ is the predicted probability of the positive class.
\underline{In PG}, the primary goal is to develop a policy $\pi_{\theta}(a|\mathbf{s})$ that guides the agent in performing actions $a$ based on the given states $\mathbf{s}$ for questions on the query set, with the aim of maximizing the cumulative reward $R$. We employ the REINFORCE algorithm~\cite{williams1992simple} to learn this policy, where $\theta$ represents the parameters of the RL-Agent. The loss function is defined as: 

\begin{equation}
\label{eq:policy}
\mathcal{L}_\text{PG} = -R\ln\pi_{\theta}(a|\mathbf{s}).
  \end{equation}

\subsection{ML-Agent: Generating Prompts for Planning}
\label{sec:mlagent}
In the ML-Agent, the most relevant path identified by the RL-Agent is selected and integrated into the prompt for a LLM. We manage transferability through the ML-Agent using Meta-Learning (ML). The rationale is that the agent is trained to structure the prompt as an in-context learning (ICL) instance by pre-appending the exemplar planning path, which can potentially improve LLM generalization to new tasks by updating with only a few examples, as evidenced in~\cite{min2021metaicl,sinha2024maml}. Below, we discuss the model architecture and training details for the ML-Agent. 

\smallskip
\noindent \textbf{Model Architecture.} As shown in Figure~\ref{fig:overall}(c), our ML-Agent uses the text encoder structure from \cite{radford2021learning} for both the question encoder and the path encoder. It employs two transformer modules with shared self-attention layers to capture potential features. We treat the instruction path and question as two text sequences ending with \texttt{[EOS]} tokens, and derive their feature representations from the activations of the highest transformer layer at these \texttt{[EOS]} tokens. The ML-Agent is trained to align the question and instruction path representations, and the most relevant path is retrieved based on the cosine similarities of these representations. Notably, the model does not use a $K$-classifier for path selection, ensuring that the architecture remains independent of the $K$ hyperparameter and does not require retraining when $K$ is adjusted.

\smallskip
\noindent \textbf{Training ML-Agent.} The ML-Agent training involves two phases: pre-training (PT) and fine-tuning (FT). \underline{In PT}, we optimize the agent using two pre-training tasks: Question Path Alignment (QPA) and Question Path Matching (QPM). For QPA, the objective is to align question and path representations by bringing similar pairs closer together and pushing dissimilar pairs apart through a contrastive approach.
Specifically, we sample a batch of question-path pairs from the support set (e.g., $Q_1^1$ and $P_1^1$ as shown in Figure~\ref{fig:overall}). For each pair, denoted by $<Q_i^j, P_i^j>$, where $Q_i^j \in \mathcal{Q}$ and $P_i^j \in \mathcal{P}$, we obtain their embedding vectors $\mathbf{v}_{i,j}^Q$ and $\mathbf{v}_{i,j}^P$ via the two encoders. We treat $\mathbf{v}_{i,j}^P$ as the positive example for $\mathbf{v}_{i,j}^Q$ (the anchor), since $Q_i^j$ and $P_i^j$ are paired, while the other paths in the batch are considered as negatives. The contrastive loss, denoted as $\mathcal{L}_{Q,P}$, encourages the paths to align with the anchor question by comparing their positive and negative pairs, that is

\begin{align}
\label{eq:clLoss_half}
\tiny
\mathcal{L}_{Q,P} = \sum\nolimits_{ Q_i^j \in \mathcal{Q} } -\log \frac{\exp\Bigr({\mathbf{v}_{i,j}^Q \cdot \mathbf{v}_{i,j}^P/\tau}\Bigr)}{ \sum\nolimits_{ P_{i'}^{j'} \in \mathcal{P}, i'\neq i, j'\neq j} \exp\Bigr({\mathbf{v}_{i,j}^Q \cdot \mathbf{v}_{i',j'}^P /\tau}\Bigr)},
\end{align}
where $\tau$ denotes a temperature parameter. Symmetrically, we can define $\mathcal{L}_{P,Q}$ by anchoring at $\mathbf{v}_{i,j}^P$. The overall loss $\mathcal{L}_\text{QPA}$ is then defined as:
\begin{align}
    \label{eq:clLoss}
    \mathcal{L}_\text{QPA} =  (\mathcal{L}_{Q,P} + \mathcal{L}_{P,Q})/2.
\end{align}

For QPM, we align questions with paths through a binary classification task. The model predicts whether a question-path pair is a match (labeled 1) or a mismatch (labeled 0). The training objective uses binary cross-entropy loss, which is defined as follows:
\begin{align}
    \label{eq:match}
    \mathcal{L}_\text{QPM} = - y * \log(P) + (y - 1) * \log(1 - P).
\end{align}
where $y$ denotes the label and $P$ represents the predicted probability of the positive class. Finally, the ML-Agent is trained using a multi-task learning approach, with the loss function $\mathcal{L}_\text{PT}$ is defined as:

\begin{align}
    \label{eq:PTLoss}
    \mathcal{L}_\text{PT} = \mathcal{L}_\text{QPA} + \mathcal{L}_\text{QPM}.
\end{align} 

\underline{In FT}, we further fine-tune the model using questions from the query set. Specifically, for each question $\hat{Q} \in \mathcal{\hat{Q}}$, we retrieve $K$ paths using the RL-Agent. We employ a hard negative mining strategy where the $K$ retrieved paths are considered as hard negative samples, since they are relevant to the question $\hat{Q}$. Additionally, we sample paths from other questions and add them to the $K$ paths, forming a path pool denoted by $\mathcal{\hat{P}}$. The performance is then evaluated by comparing the ground truth answer $A$ with the generated answer $\hat{A}$ via a LLM for each path in $\mathcal{\hat{P}}$. Based on a specific metric $\Delta(\hat{A}, A)$, the best path denoted by $\hat{P}$ is identified as the positive example for $\hat{Q}$, and the other paths in the pool are considered as negatives. The loss function $\mathcal{L}_\text{FT}$ for the fine-tuning phase is defined as:

\begin{equation}
\begin{aligned}
\label{eq:FTLoss}
\tiny
&\mathcal{L}_\text{FT} =  (\mathcal{L}'_{Q,P} + \mathcal{L}'_{P,Q})/2\\
&\mathcal{L}'_{Q,P} = \sum\nolimits_{ \hat{Q} \in \mathcal{\hat{Q}} } -\log \frac{\exp\Bigr({\mathbf{v}^{\hat{Q}} \cdot \mathbf{v}^{\hat{P}}/\tau}\Bigr)}{ \sum\nolimits_{ \overline{P} \in \mathcal{\hat{P}}, \overline{P} \neq \hat{P}} \exp\Bigr({\mathbf{v}^{\hat{Q}} \cdot \mathbf{v}^{\overline{P}} /\tau}\Bigr)},\\
\end{aligned}
\end{equation}
where $\mathbf{v}^{\hat{Q}}$ and $\mathbf{v}^{\hat{P}}$ denote the embedding vectors for $\hat{Q}$ and $\hat{P}$, respectively. $\mathcal{L}'_{P,Q}$ is a symmetric definition based on $\mathcal{L}'_{Q,P}$.

\smallskip
\noindent \textbf{Prompt Structure for LLM Generation.} The path $\hat{P}$ returned by ML-Agent is used to construct a prompt that guides the LLM in generating an answer, denoted as $\hat{A}$. Our prompt is composed of four parts, as illustrated in Figure~\ref{fig:overall}(c). (1) Task Description: This part introduces the task, detailing the specific question $\hat{Q}$ to be solved. (2) Instruction Definitions: This part provides definitions for each instruction, such as \texttt{Search[topic]} or \texttt{Lookup[entity]}. (3) Planning Path: The path $\hat{P}$ is integrated to create a structured plan, guiding the LLM through step-by-step actions to address $\hat{Q}$. (4) Demonstrations: Examples of planning paths are provided to offer reference and context for the LLM. Additionally, the \texttt{InstructRAG} framework supports integration with both trainable LLMs (e.g., fine-tuning GLM-4~\cite{glm2024chatglm} with the ground truth paths following~\cite{zhu2024knowagent}), and frozen LLMs (e.g., GPT-4o mini~\cite{achiam2023gpt} and DeepSeek-V2~\cite{deepseekv2}) to leverage its inherent capabilities for planning.

\SetKwInOut{KwIn}{Require}
\begin{algorithm}
    \small
    \caption{The \texttt{InstructRAG} - Training Stage}
    \label{alg:train}
	\KwIn{
        a training support set $\mathbb{S}$; a training query set $\mathbb{Q}$}
        randomly initialize $\theta$ for RL-Agent and $\eta$ for ML-Agent\\
        construct the instruction graph $G$ with $\mathbb{S}$ by Algorithm~\ref{alg:ig}\\
        \While {not done} {
        sample a batch of tasks $\mathcal{T}$\\
        \For{each $T_i \in \mathcal{T} (1 \le i \le |\mathcal{T}|$)} {
            evaluate $\nabla_\theta \mathcal{L}_\text{WS}^{T_i} (\text{RL-Agent}_{\theta})$ by Eq~\ref{eq:ws} wrt $\mathcal{B}$ questions for $T_i$ in $\mathbb{S}$\\
            compute adapted $\theta'_i \leftarrow \theta-\alpha\nabla_\theta \mathcal{L}_\text{WS}^{T_i} (\text{RL-Agent}_{\theta})$\\
            evaluate $\nabla_\theta \mathcal{L}_\text{PT}^{T_i} (\text{ML-Agent}_{\eta})$ by Eq~\ref{eq:PTLoss} wrt $\mathcal{B}$ questions for $T_i$ in $\mathbb{S}$ \\
            compute adapted $\eta'_i \leftarrow \eta-\alpha\nabla_\theta \mathcal{L}_\text{PT}^{T_i}(\text{ML-Agent}_{\eta})$\\
        }
        update $\theta \leftarrow \theta - \beta \nabla_\theta \sum_{T_i} \mathcal{L}_\text{PG}^{T_i} (\text{RL-Agent}_{\theta'_i})$ by Eq~\ref{eq:policy} wrt questions for all sampled tasks in $\mathbb{Q}$ \\
        update $\eta \leftarrow \eta - \beta \nabla_\eta \sum_{T_i} \mathcal{L}_\text{FT}^{T_i} (\text{ML-Agent}_{\eta'_i})$ by Eq~\ref{eq:FTLoss} wrt questions for all sampled tasks in $\mathbb{Q}$ \\
    }
\textbf{Return} trained $\text{RL-Agent}_{\theta}$ and $\text{ML-Agent}_{\eta}$
\end{algorithm}

\SetKwInOut{KwIn}{Require}
\begin{algorithm}
    \small
    \caption{The \texttt{InstructRAG} - Few-Shot Learning Stage}
    \label{alg:fewshot}
	\KwIn{
        a testing support set $\mathbb{S}'$; $\text{RL-Agent}_{\theta}$; $\text{ML-Agent}_{\eta}$}
        insert $\mathbb{S}'$ into $G$ by Algorithm~\ref{alg:ig}, and obtain $G'$\\ 
        \For{each $T_i \in \mathbb{S}' (1 \le i \le |\mathbb{S}'|$)} {
        
        $\theta'_i \leftarrow \theta-\alpha\nabla_\theta \mathcal{L}_\text{WS}^{T_i} (\text{RL-Agent}_{\theta}) - \beta\nabla_\theta \mathcal{L}_\text{PG}^{T_i} (\text{RL-Agent}_{\theta})$ by Eq~\ref{eq:ws} and Eq~\ref{eq:policy} wrt $\mathcal{B}$ questions for $T_i$ in $\mathbb{S}'$ \\
        
        $\eta'_i \leftarrow \eta-\alpha\nabla_\eta \mathcal{L}_\text{PT}^{T_i} (\text{ML-Agent}_{\eta}) - \beta\nabla_\eta \mathcal{L}_\text{FT}^{T_i} (\text{ML-Agent}_{\eta})$ by Eq~\ref{eq:PTLoss} and Eq~\ref{eq:FTLoss} wrt $\mathcal{B}$ questions for $T_i$ in $\mathbb{S}'$ \\
        }
    
\textbf{Return} adapted $\text{RL-Agent}_{\theta'_i}$ and $\text{ML-Agent}_{\eta'_i}$ for each task
\end{algorithm}

\SetKwInOut{KwIn}{Require}
\begin{algorithm}
    \small
    \caption{The \texttt{InstructRAG} - Testing Stage}
    \label{alg:test}
	\KwIn{a testing query set $\mathbb{Q}'$; $\text{RL-Agent}_{\theta'_i}$; $\text{ML-Agent}_{\eta'_i}$}
    \For{each $T_i \in \mathbb{Q}' (1 \le i \le |\mathbb{Q}'|$)} { 
       run $\text{RL-Agent}_{\theta'_i}$ and $\text{ML-Agent}_{\eta'_i}$ for questions in $T_i$\\
       evaluate the effectiveness with a metric $\Delta(\cdot,\cdot)$
    }
\textbf{Return} the average effectiveness across all tasks
\end{algorithm}

\subsection{The \texttt{InstructRAG} Framework}
\label{sec:instructrag}
We present the \texttt{InstructRAG} framework in three stages: (1) the Training Stage, (2) the Few-Shot Learning Stage, and (3) the Testing Stage. In (1), the framework employs a meta-learning approach~\cite{finn2017model} to collaboratively train two agents using both support and query sets from seen tasks. In (2), the agents' parameters are quickly adapted to unseen tasks using few-shot examples on the support set. In (3), the effectiveness of the adaptation is evaluated using the query set on these unseen tasks. 

\smallskip
\noindent\textbf{Training Stage.} 
As shown in Algorithm~\ref{alg:train}, the process inputs a support set and a query set from the seen training tasks and outputs the trained RL-Agent and ML-Agent. The support set is used to construct the instruction graph $G$ as detailed in Algorithm~\ref{alg:ig}. The two agents are then trained iteratively.
In each iteration, the RL-Agent and ML-Agent are represented as $\text{RL-Agent}_{\theta}$ and $\text{ML-Agent}_{\eta}$ with parameters $\theta$ and $\eta$, respectively. When adapting to a new task $T_i$, the parameters $\theta$ and $\eta$ are updated to $\theta'$ and $\eta'$ using Equations~\ref{eq:ws} and \ref{eq:PTLoss} based on the support set ($\alpha$ denotes a learning rate). The updated parameters are quickly computed using one or more gradient descent updates with $\mathcal{B}$ questions.
Following this, the model parameters are optimized by improving the performance of $\text{RL-Agent}_{\theta'_i}$ using Equation~\ref{eq:policy} and $\text{ML-Agent}_{\eta'_i}$ using Equation~\ref{eq:FTLoss}, with respect to $\theta$ and $\eta$ across sampled tasks from the query set ($\beta$ denotes a learning rate). Our training approach aims to optimize both agents so that a minimal number of gradient steps on a new task will produce the most effective behavior for that task.

\smallskip
\noindent\textbf{Few-shot Learning Stage.}
As shown in Algorithm~\ref{alg:fewshot}, the process adapts the trained $\text{RL-Agent}_{\theta}$ and $\text{ML-Agent}_{\eta}$ to separate models, denoted as $\text{RL-Agent}_{\theta'_i}$ and $\text{ML-Agent}_{\eta'_i}$, for each task $T_i$. This adaptation involves extending the graph $G$ to $G'$ using Algorithm~\ref{alg:ig} on the testing support set $\mathcal{S}'$. For each task, gradient descent is performed to adapt $\text{RL-Agent}_{\theta}$ (by Equations~\ref{eq:ws} and \ref{eq:policy}) and $\text{ML-Agent}_{\eta}$ (by Equations~\ref{eq:PTLoss} and \ref{eq:FTLoss}) with few-shot questions from $\mathcal{S}'$.

\smallskip
\noindent\textbf{Testing Stage.}
As shown in Algorithm~\ref{alg:test}, each task $T_i$ is performed using the corresponding adapted models ($\text{RL-Agent}_{\theta'_i}$ and $\text{ML-Agent}_{\eta'_i}$) on the testing query set $\mathcal{Q}'$. The average effectiveness, evaluated using a specific metric $\Delta(\cdot,\cdot)$, is reported across all tasks.

%% file: setup.tex
\section{EXPERIMENTS}
\label{sec:experiment}

\subsection{Experimental Setup}
\label{sec:setup}

\noindent\textbf{Datasets.} In line with previous research~\cite{zhu2024knowagent,qiao2024agent}, we conduct experiments on four widely-used task planning datasets: HotpotQA~\cite{yang2018hotpotqa}, ALFWorld~\cite{shridhar2020alfworld}, Webshop~\cite{yao2022webshop}, and ScienceWorld~\cite{wang2022scienceworld}. HotpotQA is designed for multi-hop reasoning tasks, consists of approximately 113K QA pairs sourced from Wikipedia. 
ALFWorld enables agents to complete embodied tasks in a simulated environment (e.g., placing a washed apple in the kitchen fridge). Webshop is a web application that simulates an online shopping environment, where an agent navigates webpages to find, customize, and purchase an item based on a text instruction specifying the product requirements. ScienceWorld assesses agents' scientific reasoning abilities at the level of an elementary school science curriculum.

To set up the meta-learning, for HotpotQA, we define tasks using the 12 answer types in the dataset (e.g., Person, Location, Date), where we randomly select 6 types as the seen training tasks and 6 types as the unseen testing tasks. For ALFWorld, we use their provided seen tasks and unseen tasks for training and testing, respectively. For Webshop, we define tasks by product category, where we randomly sample 60\% categories for training and the remaining for testing. For ScienceWorld, we utilize it to evaluate the generalization capability of \texttt{InstructRAG} across datasets, focusing on tasks that are entirely new to a trained \texttt{InstructRAG} model.

\smallskip\noindent\textbf{Baselines.} We carefully review the literature and identify the following baseline methods: ReAct~\cite{yao2022react}, WKM~\cite{qiao2024agent}, Reflexion~\cite{shinn2024reflexion}, GenGround~\cite{shi2024generate}, and RAP~\cite{kagaya2024rap}. These correspond to recent representative techniques discussed in Section~\ref{sec:related}. 
For GenGround, we employ a retriever implemented by LlamaIndex~\cite{Liu_LlamaIndex_2022} to find information that grounds LLM-generated answers, 
where we store TAO triplets from previous successful experiences across the datasets and utilize the retrieved similar triplets as the retriever's output.
The same data (i.e., support sets from both training and testing tasks) is used to prepare the external database for the retrieval-based methods (i.e., GenGround and RAP). 
In addition, we incorporate the \texttt{InstructRAG} and baselines into three typical LLMs, namely GLM-4~\cite{glm2024chatglm}, GPT-4o mini~\cite{achiam2023gpt}, and DeepSeek-V2~\cite{deepseekv2} for comparison.

To ensure \underline{fair performance comparisons}, we note that: 1) Both the baselines and \texttt{InstructRAG} are configured with same setups, including the same retrievers and backbone LLMs; 2) we follow the hyperparameter settings specified in their original papers.

\smallskip\noindent\textbf{Evaluation Metrics.} Following~\cite{zhu2024knowagent, qiao2024agent, liu2023bolaa}, we evaluate the effectiveness of \texttt{InstructRAG} on four datasets. For HotPotQA, the F1 score is used, comparing the agent's answers with the ground truth. For ALFWorld, the success rate, a binary metric (0 or 1), indicates whether the agent successfully completed the task. For WebShop and ScienceWorld, a reward score ranging from 0 to 1 is employed to measure the level of task completion. Overall, higher values indicate superior results. We note that all reported experimental results are \underline{statistically significant}, verified by a t-test with $p < 0.05$.

\smallskip
\noindent\textbf{Implementation Details.} 
We implement \texttt{InstructRAG} and baselines using Python 3.7. 
The threshold $\delta$ for constructing instruction graphs is set to {\Comment 0.4}. In RL-Agent, we implement a {\Comment two}-layered feedforward neural network. The first layer consists of {\Comment 20} neurons using the {\Comment tanh} activation function, and the second layer comprises 2 neurons corresponding to the action space to include or exclude a node. We use the Adam stochastic gradient descent with a learning rate of 0.001 to optimize the policy, and the reward discount is set to 0.99. In ML-Agent, the hyperparameter $K$ for selecting a path is empirically set to {\Comment 3}. 
%
To boost training efficiency, we cache the inputs and outputs generated by the LLMs during training. 


%% file: experiments.tex
\subsection{Experimental Results}
\label{sec:result}

\noindent\textbf{(1) Effectiveness Evaluation (comparison with baseline methods).} We evaluate the effectiveness of \texttt{InstructRAG} against baselines across three LLMs on unseen tasks in Table~\ref{tab:effectiveness}, \texttt{InstructRAG} consistently outperforms the baselines, demonstrating superior effectiveness. Notably, it achieves improvements of {\Comment 19.2\%}, {\Comment 9.3\%}, and {\Comment 6.1\%} over the best baseline (RAP) on HotpotQA, ALFWorld, and Webshop, respectively. This improvement can be attributed to two factors: 1) \texttt{InstructRAG} employs a graph-based organization of instruction paths, enabling the combination into new paths for more effective planning, rather than independently storing them in an external database as RAP does, and 2) it utilizes a meta-learning approach to efficiently adapt the trained model to diverse tasks.

\begin{table}[]
\caption{Effectiveness of \texttt{InstructRAG} on unseen tasks.}
\vspace*{-3mm}
\centering
\small
\setlength{\tabcolsep}{3pt}
\label{tab:effectiveness}
\begin{tabular}{llccc}
\hline
\multicolumn{1}{l}{\multirow{1}{*}{\textbf{Backbone}}} & \multicolumn{1}{l}{\multirow{1}{*}{\textbf{Method}}} & \multicolumn{1}{c}{\textbf{HotpotQA}} & \multicolumn{1}{c}{\multirow{1}{*}{\textbf{ALFWorld}}} & \multicolumn{1}{c}{\multirow{1}{*}{\textbf{Webshop}}} 
\\ \hline
\multirow{6}{*}{GLM-4} & ReAct~\cite{yao2022react} & 24.04 & 47.01 & 62.13 \\
& WKM~\cite{qiao2024agent} & - & \color{black}{64.18} & \color{black}{68.14} \\
& Reflexion~\cite{shinn2024reflexion} & 26.88 & 52.99 & 67.91 \\
& RAP~\cite{kagaya2024rap} & 27.86 & 64.18 & 69.45 \\
& GenGround~\cite{shi2024generate} & 26.97 & 58.96 & 63.18 \\ \cline{2-5}
& \texttt{InstructRAG} & \textbf{33.61} & \textbf{72.39} & \textbf{\color{black}{71.25}} \\ \hline
\multirow{6}{*}{GPT-4o mini} & ReAct~\cite{yao2022react} & 25.45 & 42.54 & 49.16 \\
& WKM~\cite{qiao2024agent} & - & \color{black}{55.22} & \color{black}{54.56} \\
& Reflexion~\cite{shinn2024reflexion} & 27.39 & 50.75 & 51.31 \\
& RAP~\cite{kagaya2024rap} & 27.66 & 56.71 & 56.31 \\
& GenGround~\cite{shi2024generate} & 28.99 & 44.03 & 53.73 \\ \cline{2-5}
& \texttt{InstructRAG} & \textbf{31.05} & \textbf{58.21} & \textbf{64.18} \\ \hline
\multirow{6}{*}{DeepSeek-V2} & ReAct~\cite{yao2022react} & 25.35 & 52.24 & 57.58 \\
& WKM~\cite{qiao2024agent} & - & \color{black}{72.39} & \color{black}{67.46} \\
& Reflexion~\cite{shinn2024reflexion} & 28.69 & 67.16 & 61.13 \\
& RAP~\cite{kagaya2024rap} & 29.82 & 72.39 & 72.72 \\
& GenGround~\cite{shi2024generate} & 33.50 & 67.16 & 62.24 \\ \cline{2-5}
& \texttt{InstructRAG} & \textbf{37.17} & \textbf{81.34} & \textbf{74.00} \\ \hline
\end{tabular}
\vspace*{-2mm}
\end{table}

\begin{table}[]
\caption{Effectiveness of \texttt{InstructRAG} across datasets (Training: HotpotQA, Testing: ScienceWorld).}
\vspace*{-3mm}
\setlength{\tabcolsep}{2.5pt}
\label{tab:across}
\begin{tabular}{l|ccc}
\hline
HotpotQA $\rightarrow$ ScienceWorld & GLM-4 & GPT-4o mini & DeepSeek-V2 \\ \hline
RAP          & 24.37 & 23.49       & 32.15       \\
\texttt{InstructRAG}  & \textbf{26.85} & \textbf{25.10}       & \textbf{33.96}       \\ \hline
\end{tabular}
\vspace*{-3mm}
\end{table}

\begin{table}[]
\caption{Effectiveness of \texttt{InstructRAG} on seen tasks.}
\vspace*{-3mm}
\centering
\small
\setlength{\tabcolsep}{3pt}
\label{tab:seen_effectiveness}
\begin{tabular}{llccc}
\hline
\multicolumn{1}{l}{\multirow{1}{*}{\textbf{Backbone}}} & \multicolumn{1}{l}{\multirow{1}{*}{\textbf{Method}}} & \multicolumn{1}{c}{\textbf{HotpotQA}} & \multicolumn{1}{c}{\multirow{1}{*}{\textbf{ALFWorld}}} & \multicolumn{1}{c}{\multirow{1}{*}{\textbf{Webshop}}} \\ 
\hline
\multirow{2}{*}{GLM-4} & RAP~\cite{kagaya2024rap} & \color{black}{27.27} & \color{black}{63.33} & \color{black}{68.35} \\
& \texttt{InstructRAG} & \textbf{34.99} & \textbf{72.50} & \textbf{\color{black}{73.60}} \\ \hline
\multirow{2}{*}{GPT-4o mini} & RAP~\cite{kagaya2024rap} & 27.18 & 60.00 & 57.92 \\
& \texttt{InstructRAG} & \textbf{31.09} & \textbf{64.17} & \textbf{64.92} \\ \hline
\multirow{2}{*}{DeepSeek-V2} & RAP~\cite{kagaya2024rap} & 31.53 & 75.83 & 71.01 \\
& \texttt{InstructRAG} & \textbf{38.81} & \textbf{84.17} & \textbf{79.17} \\ \hline
\end{tabular}
\vspace*{-2mm}
\end{table}

\begin{table}[]
\caption{Robustness to erroneous historical paths.}
\vspace*{-3mm}
\centering
\label{tab:noise}
\begin{tabular}{l|cccccc}
\hline
Noise rate  & 0\%   & 10\%  & 20\%  & 30\%  & 40\%  & 50\%  \\ \hline
RAP~\cite{kagaya2024rap}         & 29.82 & 28.74 & 27.42 & 26.01 & 24.10 & 21.72 \\
\texttt{InstructRAG} & 37.17 & 36.64 & 36.17 & 35.45 & 34.29 & 33.04\\ \hline
\end{tabular}
\vspace*{-2mm}
\end{table}  




\begin{figure*}[t]
	\centering
	\begin{tabular}{c c c c}        
		\begin{minipage}{0.24\linewidth}
			\includegraphics[width=\linewidth, trim={4.5mm 0mm 15mm 12.5mm}, clip]{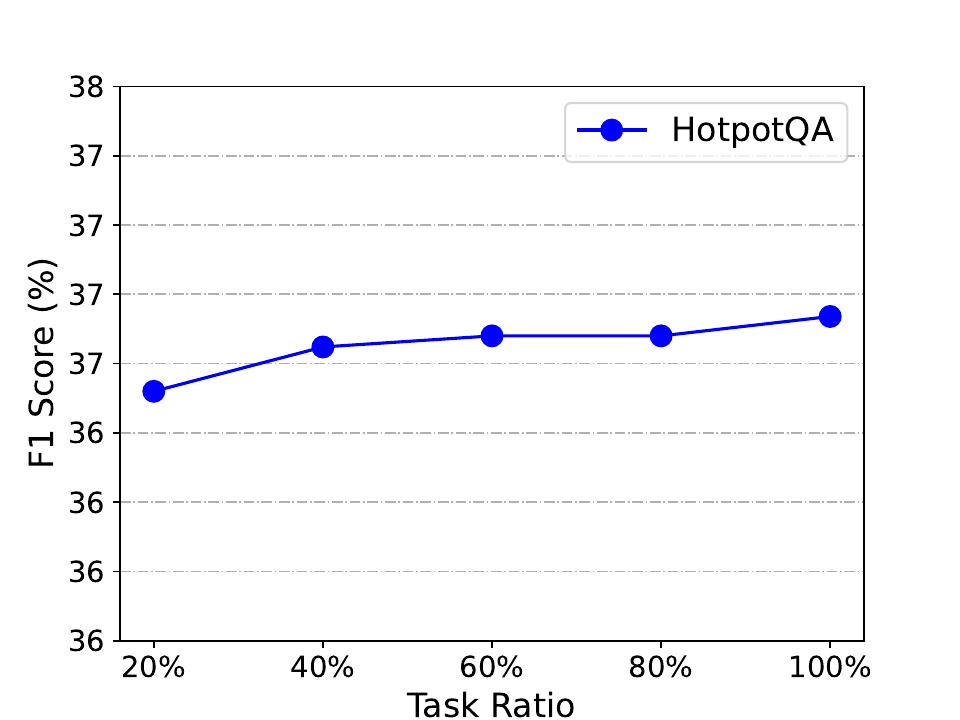}
		\end{minipage}\hspace{-3mm}
		&
		\begin{minipage}{0.24\linewidth}
			\includegraphics[width=\linewidth, trim={1.5mm 0mm 15mm 12.5mm}, clip]{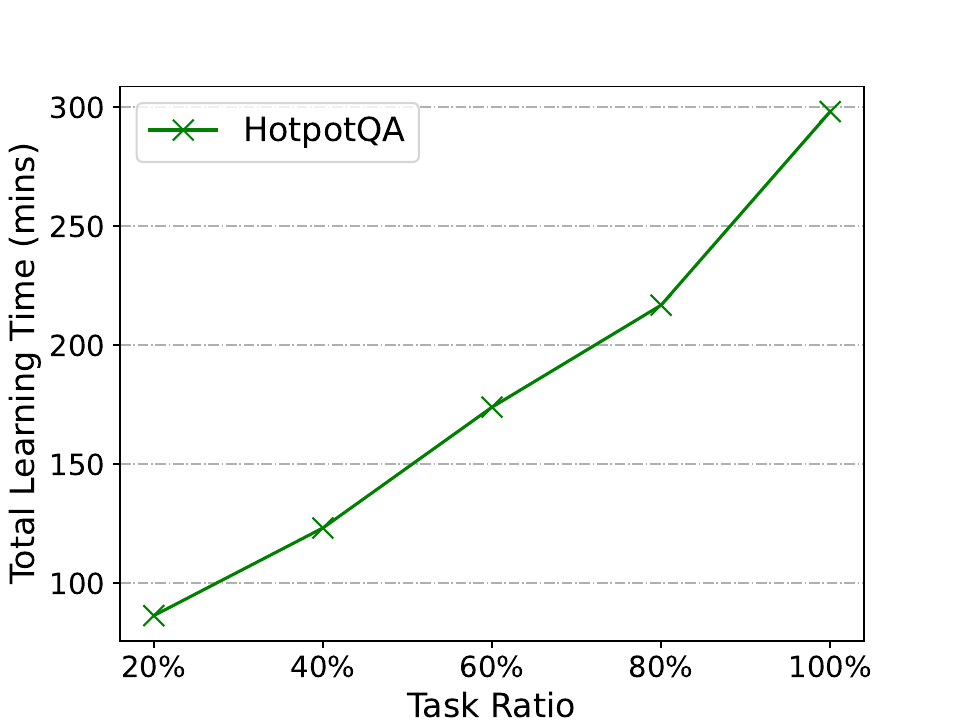}
		\end{minipage}\hspace{-3mm}
		&
		\begin{minipage}{0.24\linewidth}
			\includegraphics[width=\linewidth, trim={4.5mm 0mm 15mm 12.5mm}, clip]{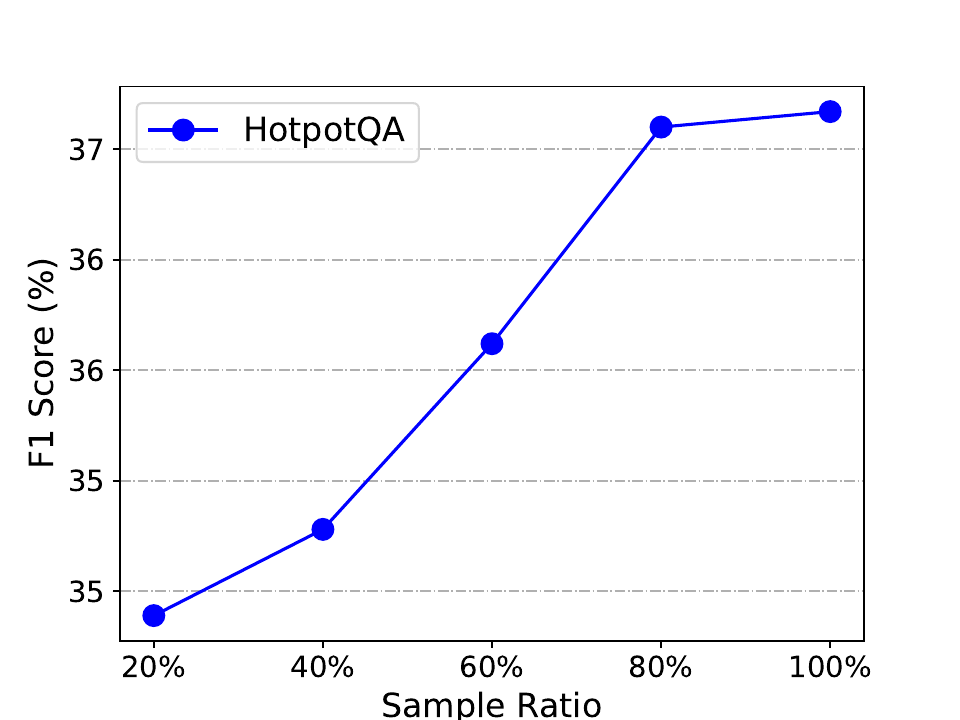}
		\end{minipage}\hspace{-3mm}
		&
		\begin{minipage}{0.24\linewidth}
			\includegraphics[width=\linewidth, trim={1.5mm 0mm 15mm 12.5mm}, clip]{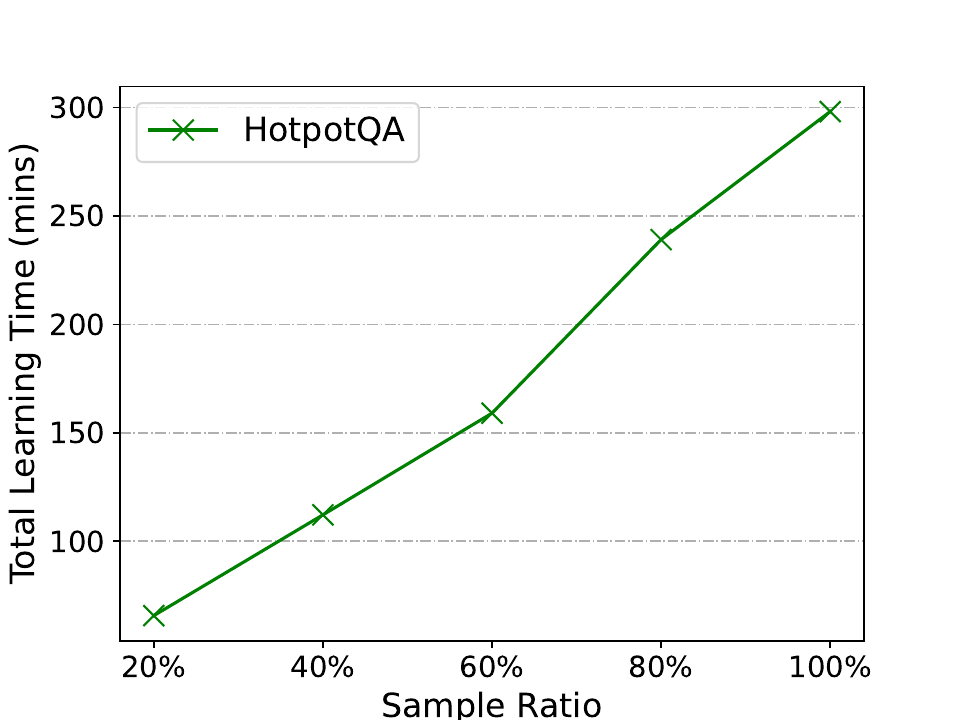}
		\end{minipage}\hspace{-3mm}
		\\
		(a) F1 Score (\#Unseen tasks) 
		&
		(b) Time (\#Unseen tasks) 
		&
		(c) F1 Score (\#Samples)
  	    &
		(d) Time (\#Samples)
	\end{tabular}
 	\vspace*{-2mm}
	\caption{F1 scores and few-shot learning times wrt the number of unseen tasks or samples with DeepSeek-V2 on HotpotQA.}
    \label{fig:increase}
    \vspace*{-2mm}
\end{figure*}

\noindent\textbf{(2) Effectiveness Evaluation (generalization capabilities across datasets).} We further evaluate the generalization capabilities of InstructRAG, by applying the trained InstructRAG model from HotpotQA to entirely new tasks in the ScienceWorld dataset. The results are reported in Table~\ref{tab:across}. Consistently, InstructRAG outperforms the best baseline method, RAP, with 6\%-10\% improvements.

\noindent\textbf{(3) Effectiveness Evaluation (performance on seen training tasks).} We also report the performance on seen training tasks. Compared to RAP, similar improvements are observed in Table~\ref{tab:seen_effectiveness}, with average improvements of {\Comment 21.9\%}, {\Comment 10.8 \%}, and {\Comment 10.4\%} on HotpotQA, ALFWorld, and Webshop, respectively. 

\noindent\textbf{(4) Effectiveness Evaluation (robustness to noise).} We examine the impact of erroneous historical paths in the instruction graph on task performance, by introducing noisy paths (i.e., failed instruction paths from past experiences) with a controlled noise rate ranging from 0\% to 50\%. For comparison, we use RAP, the best baseline method, which also includes noisy paths in its database. The F1 score results, based on DeepSeek-V2 for HotPotQA, are reported in Table~\ref{tab:noise}. Notably, even with a noise rate of 50\%, the performance of \texttt{InstructRAG} remains relatively stable, with only a 11.1\% decrease. This robustness stems from the diverse instruction combinations that help select appropriate paths and mitigate noise effectively.

\begin{table}[t]
\caption{Ablation study for verifying enlargeability (RL-Agent) and transferability (ML-Agent) on HotpotQA.}
\vspace{-3mm}
\label{tab:ablation}
\begin{tabular}{lc}
\hline
Components & F1 score  \\ \hline
\texttt{InstructRAG} & \textbf{37.17} \\ \hline
w/o Instruction Graph & 32.87 \\ \hline
w/o RL-Agent & 33.45 \\
w/o warm-start in RL-Agent  & 34.37 \\
w/o policy gradient in RL-Agent & 36.18 \\ \hline
w/o ML-Agent & 34.78 \\
w/o pre-training in ML-Agent & 36.19 \\ 
w/o  fine-tuning in ML-Agent & 36.24 \\ \hline
\end{tabular}
\vspace*{-2mm}
\end{table}

\begin{table}[t]
\setlength{\tabcolsep}{3.4pt}
\caption{Impacts of threshold $\delta$ and runtime efficiency.}
\vspace{-3mm}
\label{tab:parameter_delta}
\begin{tabular}{l|cccccc}
\hline
$\delta$ & 0.0 & 0.2 & 0.4 & 0.6 & 0.8  & 1.0 \\ \hline
F1 score         & 34.02  &35.19 &\textbf{37.17} &36.61 &36.48 &  35.93\\
Construction (s) & 19.61 & 21.27 & \textbf{20.87} & 21.65 & 23.08 & 22.07 \\
Training (hours) &23.26 & 23.64 & \textbf{23.93} & 24.81 & 25.13 & 25.17 \\
Few-shot (mins/task) & 26.35 &26.89 &\textbf{27.10} &28.01	&28.47 &28.51 \\
Testing (s) &33.87	&34.46	&\textbf{34.74}	&35.85	&36.45	&36.47 \\
\# of nodes      & 5 & 29 & \textbf{286} & 666 &720  &  725 \\
\hline
\end{tabular}
\vspace{-2mm}
\end{table}

\begin{table}[t]
\setlength{\tabcolsep}{7pt}
\caption{Impacts of the number of retrieved $K$ paths.}
\vspace{-3mm}
\label{tab:parameter_k}
\begin{tabular}{l|ccccc}
\hline
$K$            &1 & 2 & 3 & 4 & 5 \\ \hline
F1 score       & 34.78  & 36.16  & \textbf{37.17} & {36.98} & 36.77  \\
Testing (s)  &  32.05  & 33.57 & \textbf{34.74} & 35.31 & 42.09   \\
\hline
\end{tabular}
\vspace{-2mm}
\end{table}

\smallskip\noindent\textbf{(5) Ablation Study.} We perform an ablation study to assess the contributions of different components within \texttt{InstructRAG} in Table~\ref{tab:ablation}. We evaluate the following modifications: (1) omitting the instruction graph and allowing \texttt{InstructRAG} to retrieve relevant paths directly from stored individual paths; (2) omitting the RL-Agent and using a threshold-based method to determine node inclusion or exclusion, (3) the warm-start stage, (4) the policy gradient stage; (5) omitting the ML-Agent and relying solely on the path returned by the RL-Agent for testing on unseen tasks, (6) the pre-training stage, (7) the fine-tuning stage. We observe that the knowledge in the instruction graph contribute to a significant improvement of 11.6\%, and both the RL-Agent and ML-Agent contribute to the overall improvements of 11.1\% and 6.9\%, respectively. 


\smallskip\noindent\textbf{(6) Parameter Study (threshold $\delta$ for constructing instruction graphs and runtime efficiency).} As shown in Table~\ref{tab:parameter_delta}, we vary the threshold $\delta$ from 0.0 to 1.0 to control the graph construction process. As $\delta$ increases, more nodes are created, but the construction time remains stable. This is because the total number of indexed instructions with AKNN is not sensitive to the threshold. We observe that the F1 score initially increases and then decreases as $\delta$ increases. When $\delta = 0.0$, there are few instruction node sets to manage all instructions, making it difficult to accurately identify instructions for a given question due to the large set size. Conversely, when $\delta = 1.0$, the graph reduces to individual instruction paths, losing the flexibility to combine instructions into new paths. Therefore, a moderate $\delta$ leads to the best performance. 
In addition, we present the training, few-shot learning, and testing times as $\delta$ increases. Notably, training and few-shot learning require significantly more time than the graph construction, primarily due to the higher computational demands of language generation compared to the algorithmic construction. Furthermore, the graph construction is a one-time process conducted during data pre-processing.

\smallskip\noindent\textbf{(7) Parameter Study (the number of retrieved candidate paths $K$).} We vary the number of retrieved paths $K$ from 1 to 5 and report the F1 scores and testing times in Table~\ref{tab:parameter_k}. As expected, testing time increases with larger $K$ due to the consideration of more candidate paths. We observe that overall performance converges when $K$ reaches {\Comment 3}, at which point a potentially optimal path can be retrieved from the instruction graph.

\smallskip\noindent\textbf{(8) Impact of Few-shot Learning.} \texttt{InstructRAG} includes a few-shot learning stage to quickly adapt to each task. We report its effectiveness and few-shot learning time based on the number of unseen tasks or the number of samples per task with DeepSeek-V2. As shown in Figure~\ref{fig:increase}(a)-(b) , we vary the task ratio {\Comment from 0.2 to 1.0}, and observe that the effectiveness remains stable as the number of tasks increases, indicating a strong transferability across different tasks. The running time increases with more tasks due to the inclusion of additional training data. Additionally, we vary the sample ratio {\Comment from 0.2 to 1.0} for each task. As shown in Figure~\ref{fig:increase}(c) and Figure~\ref{fig:increase}(d), we observe that the effectiveness improves and converges around {\Comment 80\%} of the samples, while the running time increases as more samples are used for training. We note that, on average, a task requires 27.1 minutes for adaptation, and different tasks can be processed in parallel. The results for GLM-4 and GPT-4o mini show similar trends and are therefore omitted for brevity.

%% file: conclusion.tex
\section{CONCLUSION}
\label{sec:conclusion}

In this paper, we conduct a systematic study on leveraging RAG for task planning and identify two critical properties: enlargability and transferability. We introduce \texttt{InstructRAG}, a novel multi-agent meta-reinforcement learning solution that integrates an instruction graph, an RL-Agent, and an ML-Agent to optimize end-to-end task planning performance. Our extensive experiments on four widely used datasets, across various LLMs demonstrate that \texttt{InstructRAG} delivers superior performance and exhibits the ability to rapidly adapt to new tasks using few-shot examples. As a future direction, we plan to extend \texttt{InstructRAG} to accommodate more tasks.

%% file: appendix2.tex
\appendix \section{Appendix}
\label{sec:appendix}

\begin{table*}[t]
\setlength{\tabcolsep}{3pt}
\caption{Overview of \texttt{InstructRAG} in three stages.}
\centering
\label{tab:stage}
\begin{tabular}{l|l|l|l}
\hline
\textbf{InstructRAG} & \textbf{Training Stage}                                                                                                                                                                                                                                                                                                                & \textbf{Few-Shot Learning Stage}                                                                                                                                                                                                                                                      & \textbf{Testing Stage}                                                                                         \\ \hline
Task                 & Seen training tasks                                                                                                                                                                                                                                                                                                                    & Unseen testing tasks                                                                                                                                                                                                                                                                  & Unseen testing tasks                                                                                           \\ \hline
Data given           & \begin{tabular}[c]{@{}l@{}}Support set and query set\end{tabular}                                                                                                                                                                                                                                                                        & Support set                                                                                                                                                                                                                                                                           & Query set                                                                                                      \\ \hline
Instruction Graph           & \begin{tabular}[c]{@{}l@{}}$$G$$ (construct with the support set)\end{tabular}                                                                                                                                                                                                                                                                        & $$G'$$ (extend the support set to $G$)                                                                                                                                                                                                                                                                           & $$G'$$                                                                                                      \\ \hline
Objective            & \begin{tabular}[c]{@{}l@{}}For each iteration:\\ \quad 1. Sample batch of tasks\\ \quad 2. Optimize RL-Agent by $\mathcal{L}_{WS}$ and ML-Agent \\ \quad by $\mathcal{L}_{PT}$ on the support set for each task \\ \quad 3. Jointly optimize RL-Agent and ML-Agent by $\mathcal{L}_{PG}$ \\ \quad and $\mathcal{L}_{FT}$ on the query sets across all sampled tasks\end{tabular} & \begin{tabular}[c]{@{}l@{}}1. Update trained RL-Agent by \\ $\mathcal{L}_{WS}$ and ML-Agent by $\mathcal{L}_{PT}$ on \\ the support set for each task\\ 2. Jointly update RL-Agent and \\ ML-Agent by $\mathcal{L}_{PG}$ and $\mathcal{L}_{FT}$ on \\the support set for each task\end{tabular} & \begin{tabular}[c]{@{}l@{}}Report the average \\ effectiveness on the \\ query sets across all tasks\end{tabular} \\ \hline
\end{tabular}
\end{table*}

\begin{table*}[t]
\caption{Prompt for Overall Plan on HotpotQA.}
\label{tab:prompt_hotpotqa}
\begin{tabular}{p{\linewidth}}
\hline
Solve a question answering task with interleaving Thought, Action, Observation steps. Thought can reason about the current situation, and Action can be three types: \\
(1) Search[entity], which searches the exact entity on Wikipedia and returns the first paragraph if it exists. If not, it will return some similar entities to search.\\
(2) Lookup[keyword], which returns the next sentence containing keyword in the current passage.\\
(3) Finish[answer], which returns the answer and finishes the task. \\ \\
Here are some examples. \\
\{Examples\}\\ \\
Here is the provided action sequence: \\
\{Instruction Path\}. \\
Assess the initial understanding of the task and adjust the approach if new insights or requirements arise during the process. \\
If an action does not yield useful information or leads to a dead end, reconsider the previous steps or switch between ``Search'' and ``Lookup'' to gather more relevant data. \\ \\
Now you have to complete the following task: \\
\{Question\}
\\ \hline
\end{tabular}
\end{table*}

\begin{table*}[t]
\caption{Prompt for Overall Plan on ALFWorld.}
\label{tab:prompt_alfworld}
\begin{tabular}{p{\linewidth}}
\hline
Interact with a household to solve a task.  The following are legal actions: go, take, clean, use, examine, look, heat, cool, open, close, toggle, put, think.
When generating an action, the first word of your response must be one of the legal actions listed above.\\ \\

Here are some examples. \\
\{Examples\} \\\\
Here is the provided action sequence: \\
\{Instruction Path\}. \\ 
Assess the initial understanding of the task and adjust the approach if new insights or requirements arise during the process. \\ \\
Now you have to complete the following task: \\
\{Question\}
\\ \hline
\end{tabular}
\end{table*}

\begin{table*}[t]
\caption{Prompt for Overall Plan on Webshop.}
\label{tab:prompt_webshop}
\begin{tabular}{p{\linewidth}}
\hline
You are an advanced reasoning agent tasked with interacting with a shopping website. The following are legal actions: \\
(1) search[keyword]: You can perform a search using specific keywords (if ``has\_search\_bar'' is True). Keep the keyword short and concise. Avoid overly detailed descriptions. Only include keywords that help identify the product. \\
(2) click[clickables]:  You can click on available clickable items. \\ \\

Here are some examples. \\
\{Examples\} \\ \\ 
Here is the provided action sequence: \\
\{Instruction Path\}.

Assess the initial understanding of the task and adjust the approach if new insights or requirements arise during the process. \\ \\
Now you have to complete the following task: \\
\{Question\}
\\ \hline
\end{tabular}
\end{table*}

\begin{figure*}[t]
	\centering
	\begin{tabular}{c c c c}
		\begin{minipage}{0.24\linewidth}
			\includegraphics[width=\linewidth, trim={4.5mm 0mm 15mm 12.5mm}, clip]{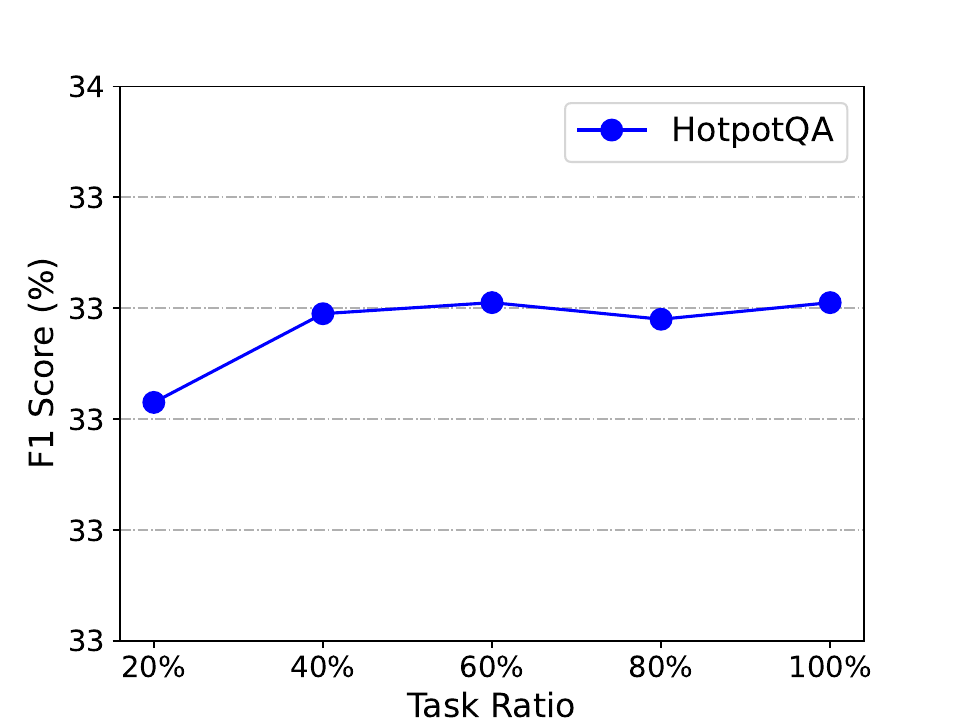}
		\end{minipage}\hspace{-3mm}
		&
		\begin{minipage}{0.24\linewidth}
			\includegraphics[width=\linewidth, trim={1.5mm 0mm 15mm 12.5mm}, clip]{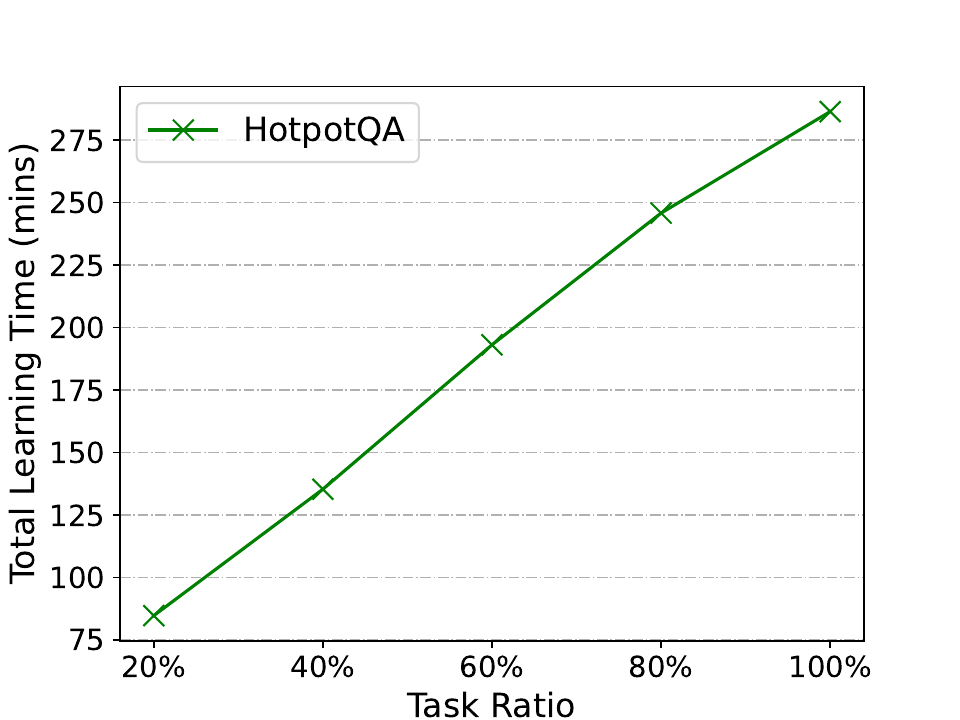}
		\end{minipage}\hspace{-3mm}
		&
		\begin{minipage}{0.24\linewidth}
			\includegraphics[width=\linewidth, trim={4.5mm 0mm 15mm 12.5mm}, clip]{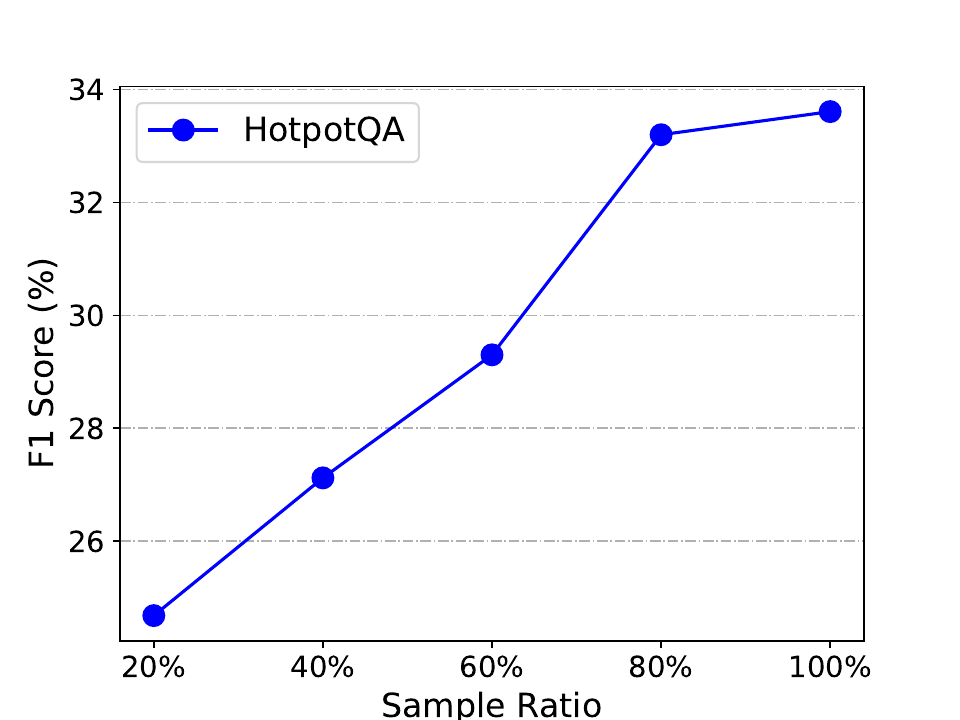}
		\end{minipage}\hspace{-3mm}
		&
		\begin{minipage}{0.24\linewidth}
			\includegraphics[width=\linewidth, trim={1.5mm 0mm 15mm 12.5mm}, clip]{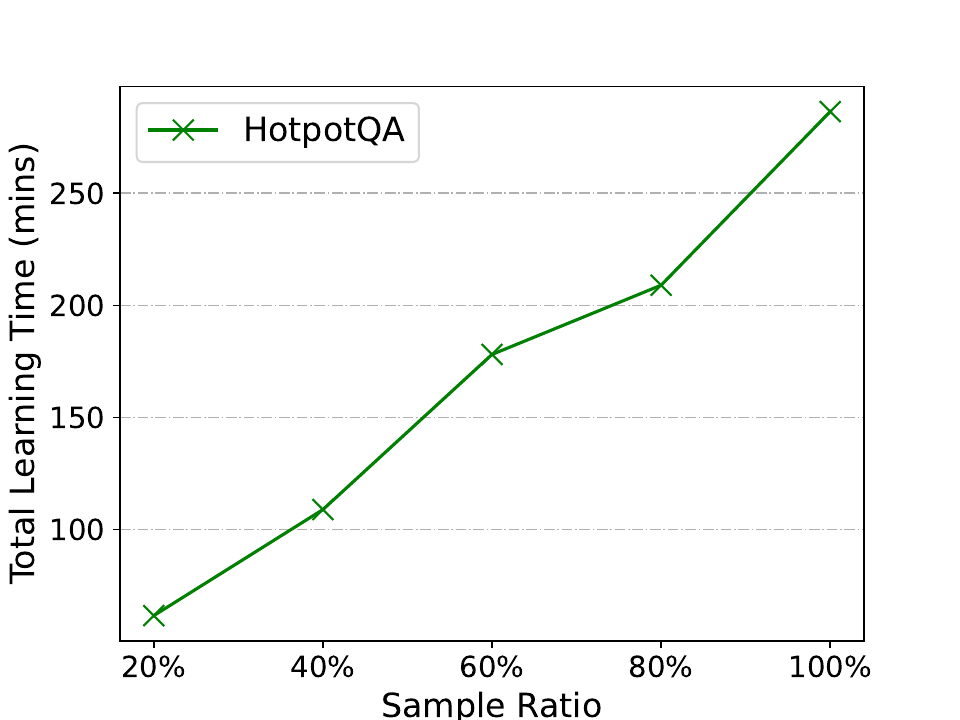}
		\end{minipage}\hspace{-3mm}
		\\
		(a) F1 Score (\#Unseen tasks) 
		&
		(b) Few-shot Time (\#Unseen tasks) 
		&
		(c) F1 Score (\#Samples)
  	&
		(d) Few-shot Time (\#Samples)
        \\
		\begin{minipage}{0.24\linewidth}
			\includegraphics[width=\linewidth, trim={4.5mm 0mm 15mm 12.5mm}, clip]{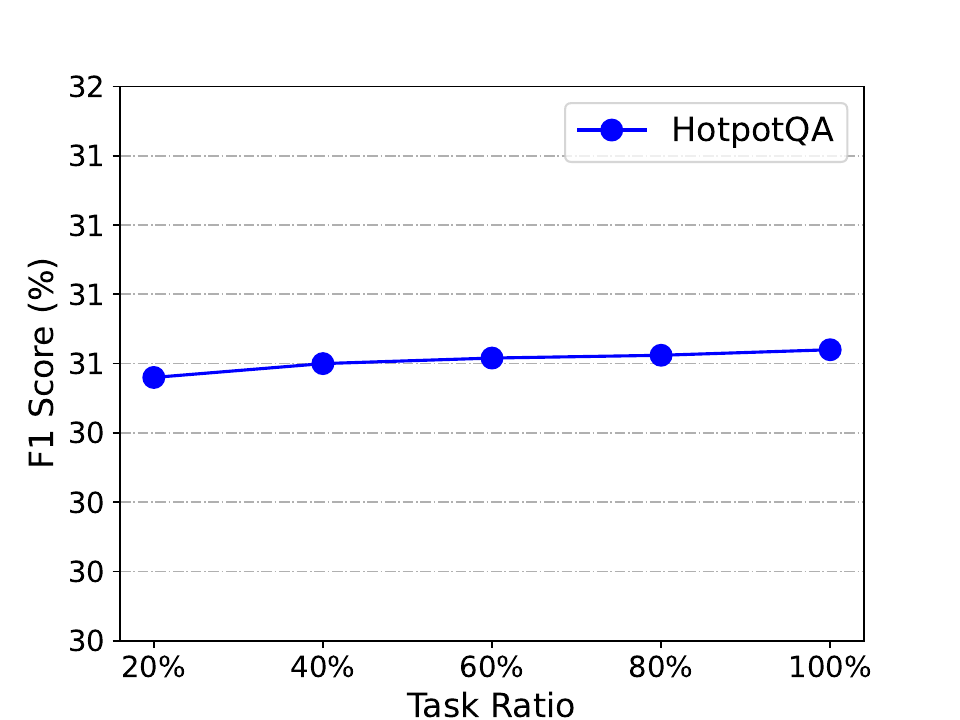}
		\end{minipage}\hspace{-3mm}
		&
		\begin{minipage}{0.24\linewidth}
			\includegraphics[width=\linewidth, trim={1.5mm 0mm 15mm 12.5mm}, clip]{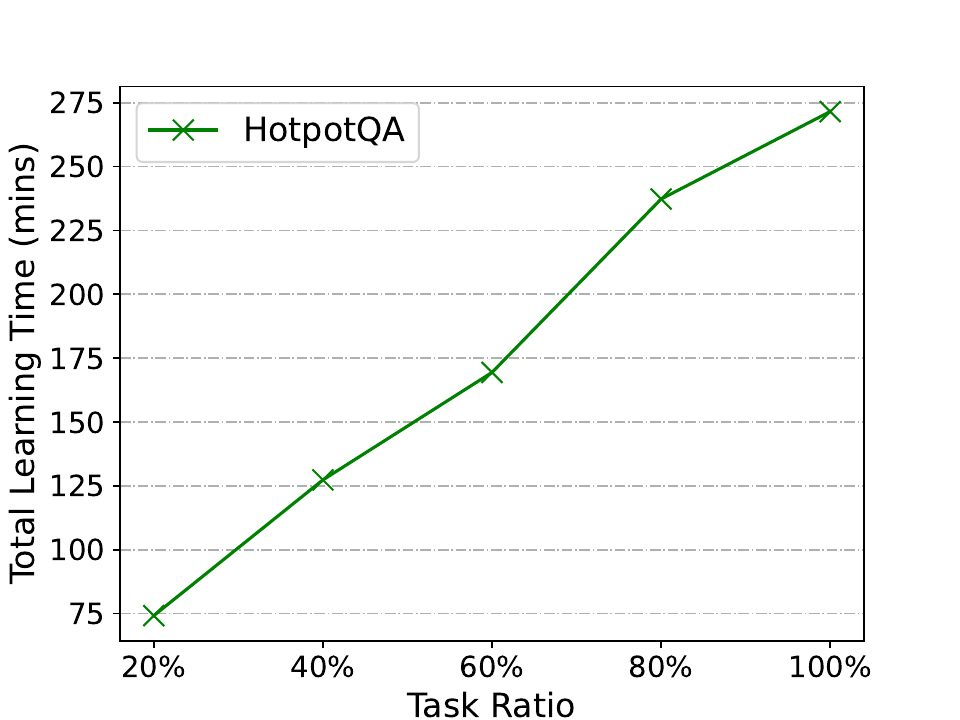}
		\end{minipage}\hspace{-3mm}
		&
		\begin{minipage}{0.24\linewidth}
			\includegraphics[width=\linewidth, trim={4.5mm 0mm 15mm 12.5mm}, clip]{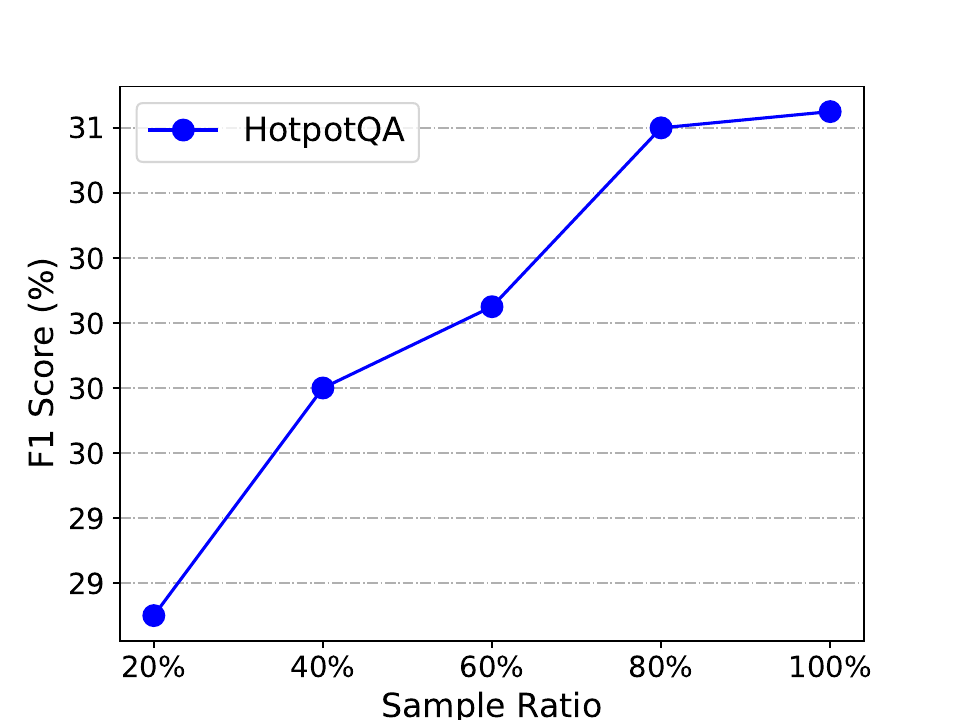}
		\end{minipage}\hspace{-3mm}
		&
		\begin{minipage}{0.24\linewidth}
			\includegraphics[width=\linewidth, trim={1.5mm 0mm 15mm 12.5mm}, clip]{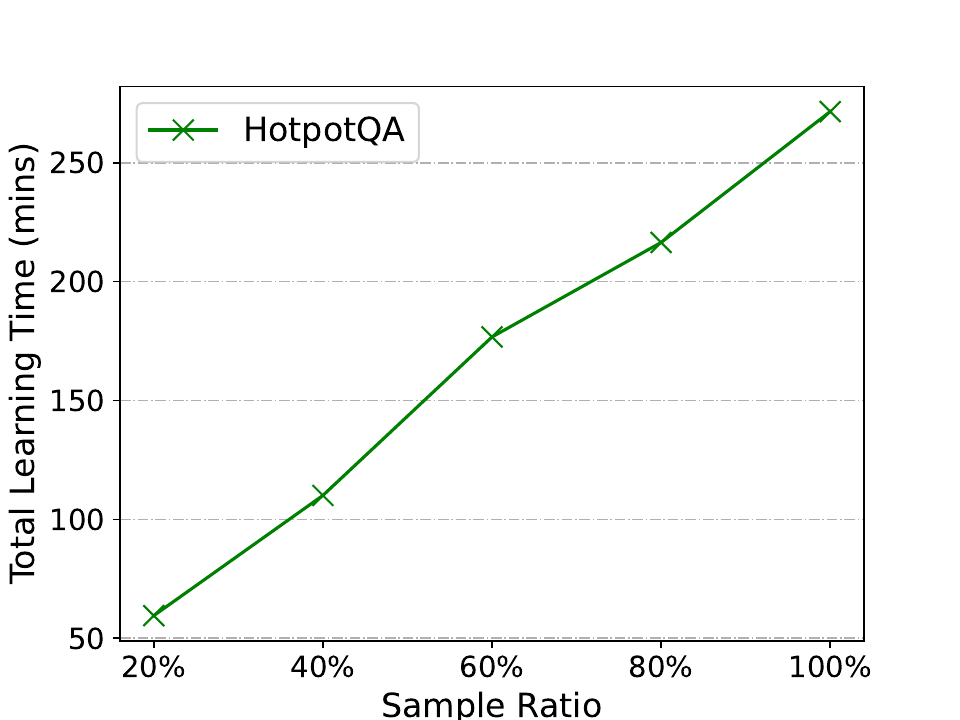}
		\end{minipage}\hspace{-3mm}
		\\
		(e) F1 Score (\#Unseen tasks) 
		&
		(f) Few-shot Time (\#Unseen tasks) 
		&
		(g) F1 Score (\#Samples)
  	&
		(h) Few-shot Time (\#Samples)
	\end{tabular}
	\caption{F1 scores and few-shot learning times wrt the number of unseen tasks or samples on HotpotQA, where (a)-(d) are for GLM-4, and (e)-(h) are for GPT-4o mini.}
    \label{fig:increase_others}
\end{figure*}

\subsection{Overview of \texttt{InstructRAG} in Three Stages}
\label{asec:stages}
The three stages of \texttt{InstructRAG}—training, few-shot learning, and testing—are summarized in Table~\ref{tab:stage}.

\subsection{Prompts}
\label{asec:prompts}
We provide the \texttt{InstructRAG} prompts for HotpotQA, ALFWorld, and Webshop in Table~\ref{tab:prompt_hotpotqa}, Table~\ref{tab:prompt_alfworld}, and Table~\ref{tab:prompt_webshop}, respectively.

\subsection{Discussion on the Use of Multi-Agent for Task Planning}
\label{asec:discussion_ma}
We provide a discussion to explain the rationale behind using both RL-Agent and ML-Agent for task planning instead of modifying a single agent (e.g., RL-Agent by setting $K=1$) to address both enlargeability and transferability. The task planning in this study requires addressing two key properties: enlargeability and transferability. These properties are somewhat orthogonal: enlargeability involves combining instructions for questions within the seen tasks, while transferability focuses on rapid adaptation to the unseen tasks. It is challenging for a single agent to optimize effectively in both directions simultaneously. Therefore, we design a multi-agent framework collaborated with two distinct agents: the RL-Agent provides candidate paths for the ML-Agent, while the ML-Agent supplies rewards for the RL-Agent. This strategic division of labor enables us to explicitly optimize for both enlargeability and transferability through multi-agent meta-reinforcement learning, and we validate the solution via an ablation study presented in Table~\ref{tab:ablation}.

\subsection{Few-shot Learning with Other LLMs}
\label{asec:few-shot}
We report the F1 scores and few-shot learning times for GLM-4 and GPT-4o mini in Figure~\ref{fig:increase_others}. Overall, similar trends can be observed, consistent with the results from DeepSeek-V2.

\subsection{Qualitative Results}
\label{asec:qualitative}
Both \texttt{InstructRAG} and RAP leverage past experiences (e.g., instruction paths) to guide LLM planning. Table~\ref{tab:traj_hotpotqa}, Table~\ref{tab:traj_alfworld}, and Table~\ref{tab:traj_webshop} illustrate the planning trajectories of \texttt{InstructRAG} and RAP for HotpotQA, ALFWorld, and Webshop, respectively. We note that \texttt{InstructRAG} combines multiple paths from related tasks into an instruction path, effectively guiding LLM planning. This is demonstrated by the overlap of several instructions (highlighted in yellow) from the instruction path in successful plans. Specifically, we analyze the planning results in Table~\ref{tab:traj_hotpotqa}. InstructRAG demonstrates an advantage by combining two paths based on a common instruction, search[Piers Haggard]. This approach effectively links two key items—Anthony Minghella, representing the novel, and Piers Haggard, representing the film adaptation of ``The Talented Mr. Ripley''. These connections enable the LLM to generate a correct query formulation, allowing it to retrieve information about the film director within the generated thoughts (e.g., Thought 5). In contrast, RAP struggles to produce a correct query based on its retrieved experiences. Its generated thoughts fail to support effective query formulation, often resulting in a planning deadlock.

\begin{table*}[t]
\caption{Comparison of \texttt{InstructRAG} and RAP with HotpotQA trajectories, where we highlight tasks, successful results, and failure results in purple, green, and red, respectively. Different colors are used to label the instructions in the instruction path, which combines Path 1 and Path 2. Junction instructions from the paths are highlighted in gray, and instructions (actions) in the planning trajectories that overlap with the instruction path are highlighted in yellow.}
\label{tab:traj_hotpotqa}
\begin{tabular}{ll}
\hline
\multicolumn{2}{p{\linewidth}}{\textbf{Task Description}: Your task is to answer a question.

\colorbox{blue!15}{%
    \parbox{0.78\linewidth}{%
        Between the directors of ``The Decline of Western Civilization'' and ``The Talented Mr. Ripley'',  who is older?
    }
}
}        \\\\

\multicolumn{1}{p{0.5\linewidth}}{\textbf{\texttt{InstructRAG}}

\underline{Instruction Path}: {\color{blue}Search[Anthony Minghella]} $\rightarrow$ {\color{blue}Search[Piers Haggard]} $\rightarrow$ {\color{orange}Search[The Decline of Western Civilization]} $\rightarrow$ {\color{orange}Search[Penelope Spheeris]} (Derived from combining tasks: ``Are both Anthony Minghella and Piers Haggard film diectors?'' and ``Which director, Piers Haggard or Penelope Spheeris, is best known as a documentary film director whose works include the trilogy titled  `The Decline of Western Civilization'?'')

\bigskip
Combined from: 

\underline{Path 1}: {\color{blue}Search[Anthony Minghella]} $\rightarrow$ {\color{blue}\colorbox{gray!15}{{Search[Piers Haggard]}}} $\rightarrow$ Finish[yes]

\smallskip
\underline{Path 2}: {\color{orange}\colorbox{gray!15} {Search[Piers Haggard]}} $\rightarrow$ {\color{orange}Search[The Decline of Western Civilization]} $\rightarrow$ {\color{orange}Search[Penelope Spheeris]} $\rightarrow$ Finish[Penelope Spheeris]

\bigskip
Thought 1: I need to search the directors of `The Decline of Western Civilization' and `The Talented Mr. Ripley', find their birthdates, and then compare them to determine who is older.

\colorbox{yellow!30}{Act 1: Search[The Decline of Western Civilization]}

Obs 1: The Decline of Western Civilization is a 1981 American documentary filmed through 1979 and 1980...

Thought 2: The director of `The Decline of Western Civilization' is Penelope Spheeris. I need to search her birthdate.

\colorbox{yellow!30}{Act 2: Search[Penelope Spheeris]}

Obs 2: Penelope Spheeris (born December 2, 1945)[1] is an American film director, producer, and screenwriter...







...

Thought 5: I need to search for the director of the 1999 film adaptation of `The Talented Mr. Ripley'. The correct search term should be `The Talented Mr. Ripley (film)'.

Act 5: Search[The Talented Mr. Ripley (film)]

Obs 5: The Talented Mr. Ripley is a 1999 American psychological thriller film written and directed by Anthony Minghella, based on Patricia Highsmith's 1955 novel of the same title...

Thought 6: The director of `The Talented Mr. Ripley' (1999 film) is Anthony Minghella. I need to search his birthdate.

\colorbox{yellow!30}{Act 6: Search[Anthony Minghella]}

Obs 6: Anthony Minghella, CBE (6 January 1954 – 18 March 2008) was a British film director, playwright, and screenwriter...

Thought 7: Anthony Minghella was born on January 6, 1954. Comparing the birthdates, December 2, 1945 (Penelope Spheeris) < January 6, 1954 (Anthony Minghella), so Penelope Spheeris is older.

Act 7: Finish[Penelope Spheeris]

Obs 7: Episode finished, Answer is CORRECT

\bigskip
\colorbox{green!15}{Success flag = True, Success rate = 1.0}} & \multicolumn{1}{p{0.5\linewidth}} {\textbf{RAP}

\underline{Paths}: Separately retrieved from the tasks: ``Which director, Piers Haggard or Penelope Spheeris, is best known as a documentary film director whose works include the trilogy titled  `The Decline of Western Civilization'?'', ``Are both Anthony Minghella and Piers Haggard film diectors?'', ``Extras, is a British sitcom about extras working in television, film, and theatre, the series was co-produced by the BBC and HBO and was created, written, and directed by Ricky Dene Gervais, an English comedian, actor, writer, producer, director, singer, and musician, born on which date?'', and ``What  teen sex comedy film written by Adam Herz and directed by brothers Paul and Chris Weitz, starred an actor best known for her role as Jessica ?''.

\bigskip
Thought 1: I need to identify the directors of `The Decline of Western Civilization' and `The Talented Mr. Ripley' to determine who is older.

Act 1: Search[The Decline of Western Civilization]

Obs 1: The Decline of Western Civilization is a 1981 American documentary filmed through 1979 and 1980...

Thought 2: Penelope Spheeris is the director of `The Decline of Western Civilization'. Next, I need to find the director of `The Talented Mr. Ripley'.

Act 2: Search[The Talented Mr. Ripley]

Obs 2: The Talented Mr. Ripley is a 1955 psychological thriller novel by Patricia Highsmith...

Thought 3: The search for `The Talented Mr. Ripley' provided information about the novel and its adaptations but did not mention the director of the 1999 film adaptation. I need to search for the director of the 1999 film `The Talented Mr. Ripley'.

Act 3: Search[director of The Talented Mr. Ripley 1999 film]

Obs 3: Could not find director of The Talented Mr. Ripley 1999 film...

Thought 4: I need to search for the director of the 1999 film `The Talented Mr. Ripley' directly by searching for the film title and identifying the director from the information provided.

Act 4: Search[The Talented Mr. Ripley 1999 film]

Obs 4: Could not find The Talented Mr. Ripley 1999 film...

Thought 5: I need to search for the director of the 1999 film `The Talented Mr. Ripley' directly by searching for the film title and identifying the director from the information provided.

Act 5: Search[The Talented Mr. Ripley 1999 film]

Obs 5: Could not find The Talented Mr. Ripley 1999 film...

Thought 6: I need to search for the director of the 1999 film `The Talented Mr. Ripley' directly by searching for the film title and identifying the director from the information provided.

Act 6: Search[The Talented Mr. Ripley 1999 film]

Obs 6: Could not find The Talented Mr. Ripley 1999 film...

Thought 7: I need to search for the director of the 1999 film `The Talented Mr. Ripley' directly by searching for the film title and identifying the director from the information provided.

Act 7: Search[The Talented Mr. Ripley 1999 film]

Obs 7: Could not find The Talented Mr. Ripley 1999 film...

\bigskip
\colorbox{red!15}{Success flag = False, Success rate = 0.0}}
\\ \hline
\end{tabular}
\end{table*}

\begin{table*}[t]
\caption{Comparison of \texttt{InstructRAG} and RAP with ALFWorld trajectories.}
\vspace{-4mm}
\label{tab:traj_alfworld}
\begin{tabular}{ll}
\hline
\multicolumn{2}{p{\linewidth}}{\textbf{Task Description}: You are in the middle of a room. Looking quickly around you, you see a cabinet 4, a cabinet 3, a cabinet 2, a cabinet 1, a countertop 1, a garbagecan 1, a handtowelholder 2, a handtowelholder 1, a sinkbasin 2, a sinkbasin 1, a toilet 1, a toiletpaperhanger 1, and a towelholder 1.

\colorbox{blue!15}{Your task is to: put two soapbar in garbagecan.}
}        \\\\
\multicolumn{1}{p{0.5\linewidth}}{\textbf{\texttt{InstructRAG}}

\underline{Instruction Path}: {\color{blue} think: to solve the task, i need to find and take a soapbar, then clean it with sinkbasin, then put it in cabinet} $\rightarrow$ {\color{blue}go to toilet 1} $\rightarrow$ {\color{blue}take soapbar 1 from toilet 1} $\rightarrow$ {\color{blue}go to sinkbasin 1} $\rightarrow$ {\color{blue}clean soapbar 1 with sinkbasin 1 $\rightarrow$} {\color{orange} go to drawer 1} $\rightarrow$  {\color{orange}open drawer 1} $\rightarrow$ {\color{cyan} take dishsponge 3 from drawer 1} $\rightarrow$ {\color{cyan}go to garbagecan 1} $\rightarrow$ {\color{cyan} put dishsponge 3 in/on garbagecan 1} (Derived from combining tasks: ``put a clean soapbar in cabinet'', ``put a clean soapbar in drawer'' and ``put two dishsponge in garbagecan'')

\bigskip
Combined from: 

\underline{Path 1}: {\color{blue} think: to solve the task, i need to find and take a soapbar, then clean it with sinkbasin, then put it in cabinet} $\rightarrow$ {\color{blue} go to toilet 1} $\rightarrow$ {\color{blue} take soapbar 1 from toilet 1} $\rightarrow$ {\color{blue} go to sinkbasin 1} $\rightarrow$ {\color{blue} \colorbox{gray!15}{clean soapbar 1 with sinkbasin 1}} $\rightarrow$ go to cabinet 1 $\rightarrow$ open cabinet 1 $\rightarrow$ put soapbar 1 in/on cabinet 1

\smallskip
\underline{Path 2}: think: to solve the task, i need to find and take a soapbar, then clean it with sinkbasin, then put it in drawer. $\rightarrow$ go to bathtubbasin 1 $\rightarrow$ take soapbar 1 from bathtubbasin 1 $\rightarrow$ go to sinkbasin 1 $\rightarrow$  {\color{orange} \colorbox{gray!15}{clean soapbar 1 with sinkbasin 1}} $\rightarrow$ {\color{orange} go to drawer 1} $\rightarrow$ {\color{orange}\colorbox{gray!15}{open drawer 1}} $\rightarrow$ put soapbar 1 in/on drawer 1

\smallskip
\underline{Path 3}: think: to solve the task, i need to find and take the first dishsponge, then put it in garbagecan, then find and take the second dishsponge, then put it in garbagecan. $\rightarrow$ go to drawer 1 $\rightarrow$  {\color{cyan} \colorbox{gray!15}{open drawer 1}} $\rightarrow$ {\color{cyan}take dishsponge 3 from drawer 1} $\rightarrow$ {\color{cyan} go to garbagecan 1} $\rightarrow$ {\color{cyan} put dishsponge 3 in/on garbagecan 1} $\rightarrow$ go to countertop 1 $\rightarrow$ take dishsponge 2 from countertop 1 $\rightarrow$ go to garbagecan 1 $\rightarrow$ put dishsponge 2 in/on garbagecan 1

\bigskip
Act 1: think: To solve the task, I need to find and take the first soapbar, then put it in garbagecan, then find and take the second soapbar, then put it in garbagecan.

Obs 1: OK.

\colorbox{yellow!30}{Act 2:  go to toilet 1}

Obs 2: On the toilet 1, you see a soapbar 2, and a soapbar 1.


...

\colorbox{yellow!30}{Act 4: take soapbar 1 from toilet 1}

Obs 4: You pick up the soapbar 1 from the toilet 1.


...

\colorbox{yellow!30}{Act 6: go to garbagecan 1}

Obs 6: On the garbagecan 1, you see nothing.

Act 7: put soapbar 1 in/on garbagecan 1

Obs 7: You put the soapbar 1 in/on the garbagecan 1.



...

\colorbox{yellow!30}{Act 9: go to toilet 1}

Obs 9: On the toilet 1, you see a soapbar 2.


...

Act 11: take soapbar 2 from toilet 1

Obs 11: You pick up the soapbar 2 from the toilet 1.


...

\colorbox{yellow!30}{Act 13: go to garbagecan 1}

Obs 13: On the garbagecan 1, you see a soapbar 1.

Act 14: put soapbar 2 in/on garbagecan 1

Obs 14: You put the soapbar 2 in/on the garbagecan 1.

\bigskip
\colorbox{green!15}{Success flag = True, Success rate = 1
}} & \multicolumn{1}{p{0.5\linewidth}} {\textbf{RAP}

\underline{Path}: Separately retrieved from the tasks: ``put a soapbottle in garbagecan'',  ``put a clean soapbar in drawer'', ``put a clean soapbar in toilet'' and ``put two dishsponge in garbagecan''

\bigskip

Act 1:think: To solve the task, I need to find and take the first soapbar, then put it in the garbagecan, then find and take the second soapbar, then put it in the garbagecan

Obs 1: OK.

Act 2: go to countertop 1

Obs 2: On the countertop 1, you see a candle 3, a candle 1, a soapbottle 1, and a spraybottle 2.

Act 3: take soapbar 1 from countertop 1

Obs 3: Nothing happens.

Act 4: go to sinkbasin 1

Obs 4: On the sinkbasin 1, you see nothing.

Act 5: go to sinkbasin 2

Obs 5: On the sinkbasin 2, you see nothing.

Act 6: go to countertop 1

Obs 6: On the countertop 1, you see a candle 3, a candle 1, a soapbottle 1, and a spraybottle 2.

Act 7: take soapbar 1 from countertop 1

Obs 7: Nothing happens.

Act 8: go to cabinet 1

Obs 8: The cabinet 1 is closed.

Act 9: open cabinet 1

Obs 9: You open the cabinet 1. The cabinet 1 is open. In it, you see a cloth 2.

Act 10: take soapbar 1 from cabinet 1

Obs 10: Nothing happens. 

...

Act 46: go to sinkbasin 1

Obs 46: On the sinkbasin 1, you see nothing.

Act 47: go to sinkbasin 2

Obs 47: On the sinkbasin 2, you see nothing.

Act 48: go to countertop 1

Obs 48: On the countertop 1, you see a candle 3, a candle 1, a soapbottle 1, and a spraybottle 2.

Act 49: take soapbar 1 from countertop 1

Obs 49: Nothing happens.

\bigskip
\colorbox{red!15}{Success flag = False, Success rate = 0}}
\\ \hline
\end{tabular}
\end{table*}

\begin{table*}[t]
\caption{Comparison of \texttt{InstructRAG} and RAP with Webshop trajectories.
}
\label{tab:traj_webshop}
\begin{tabular}{ll}
\hline
\multicolumn{2}{p{\linewidth}}{\textbf{Task Description}: Find me home office furniture sets for dining room, living room with color: navy | red, and item shape: round, and size: 21 in x 35 in, and price lower than 60.00 dollars.

\colorbox{blue!15}{%
    \parbox{\linewidth}{%
        Your task is to: look for home office furniture sets for dining room, 
        living room with color: navy | red, and item shape: round, and size: 
        21 in x 35 in, and price lower than 60.00 dollars.
    }
}
}        \\\\

\multicolumn{1}{p{0.5\linewidth}}{\textbf{\texttt{InstructRAG}}

\underline{Instruction Path}: {\color{blue} search[home office furniture sets dining room living room navy red round 1 ft 9 in x 7 ft 2 in price < 60.00]} $\rightarrow$ {\color{blue} click[b07fkgqkz1]} $\rightarrow$ {\color{orange} click[21 in x 35 in]} $\rightarrow$ {\color{orange} click[taupe | orange]} $\rightarrow$ {\color{cyan} click[round]} $\rightarrow$ {\color{cyan} click[3 ft 7 in x 5 ft 2 in]} $\rightarrow$ {\color{cyan} click[buy now]} (Derived from combining tasks: ``Find me home office furniture sets for dining room, living room with color: navy | red, and item shape: round, and size: 1 ft 9 in x 7 ft 2 in, and price lower than 60.00 dollars'', ``Find me home office furniture sets for dining room, living room with color: taupe | orange, and item shape: runner, and size: 21 in x 35 in, and price lower than 60.00 dollars'' and ``Find me home office furniture sets for dining room, living room with color: taupe | orange, and item shape: round, and size: 3 ft 7 in x 5 ft 2 in, and price lower than 60.00 dollars'' )

\bigskip
Combined from: 

\underline{Path 1}: {\color{blue} search[home office furniture sets dining room living room navy red round 1 ft 9 in x 7 ft 2 in price < 60.00]} $\rightarrow$ \colorbox{gray!15}{{\color{blue} click[b07fkgqkz1]}} $\rightarrow$ click[1 ft 9 in x 7 ft 2 in] $\rightarrow$ click[navy | red] $\rightarrow$ click[round] $\rightarrow$ click[1 ft 9 in x 7 ft 2 in] $\rightarrow$ click[buy now]

\smallskip
\underline{Path 2}: search[taupe orange runner 21 in x 35 in furniture set] $\rightarrow$ \colorbox{gray!15}{{\color{orange} click[b07fkgqkz1]}} $\rightarrow$ {\color{orange} click[21 in x 35 in]} $\rightarrow$ \colorbox{gray!15}{{\color{orange} click[taupe | orange]}} $\rightarrow$ click[runner] $\rightarrow$ click[21 in x 35 in] $\rightarrow$ click[buy now]

\smallskip
\underline{Path 3}: search[home office furniture sets dining room living room color taupe orange shape round size 3 ft 7 in x 5 ft 2 in price < 60.00] $\rightarrow$ click[b07fkgqkz1] $\rightarrow$ click[3 ft 7 in x 5 ft 2 in] $\rightarrow$ \colorbox{gray!15}{{\color{cyan} click[taupe | orange]}} $\rightarrow$ {\color{cyan} click[round] } $\rightarrow$ {\color{cyan} click[3 ft 7 in x 5 ft 2 in]} $\rightarrow$ {\color{cyan}click[buy now]}

\bigskip
Act 1: reset

Obs 1: {`has\_search\_bar': True, `clickables': [`search']}

Act 2: search[home office furniture sets dining room living room navy red round 21 in x 35 in price < 60.00]

Obs 2: {`has\_search\_bar': False, `clickables': [`back to search', `next >', `b07fkgqkz1', ...... , `b07fvhpk5x']}

\colorbox{yellow!30}{Act 3: click[b07fkgqkz1]}

Obs 3: {`has\_search\_bar': False, `clickables': [`back to search', `< prev', `description', `features', `reviews', `buy now', `1 ft 9 in x 7 ft 2 in', ...... , `21 in x 35 in', ...... , `navy-multi', `taupe']}

\colorbox{yellow!30}{Act 4: click[21 in x 35 in]}

Obs 4: {`has\_search\_bar': False, `clickables': [`back to search', `< prev', `description', `features', `reviews', `buy now', `1 ft 9 in x 7 ft 2 in', ...... , `21 in x 35 in', `runner', `rectangular', `round', `black-gray', ...... , `taupe']}

\colorbox{yellow!30}{Act 5: click[round]}

Obs 5: {`has\_search\_bar': False, `clickables': [`back to search', `< prev', `description', `features', `reviews', `buy now', `1 ft 9 in x 7 ft 2 in', ...... , `21 in x 35 in', `runner', `rectangular', `round', `black-gray', ...... , `navy | red', ...... , `taupe']}

Act 6: click[navy | red]

Obs 6: {`has\_search\_bar': False, `clickables': [`back to search', `< prev', `description', `features', `reviews', `buy now', `1 ft 9 in x 7 ft 2 in', ...... , `21 in x 35 in', `runner', `rectangular', `round', `black-gray', ...... , `taupe']}

\colorbox{yellow!30}{Act 7: click[buy now]}

\bigskip
\colorbox{green!15}{Success flag = True, Success rate = 1.0}} & \multicolumn{1}{p{0.5\linewidth}} {\textbf{RAP}

\underline{Paths}: Separately retrieved from the tasks: ``Find me home office furniture sets for dining room, living room with color: blue | gray, and item shape: runner, and size: 21 in x 35 in, and price lower than 60.00 dollars'', ``Find me home office furniture sets for dining room, living room with color: navy | red, and item shape: round, and size: 1 ft 9 in x 7 ft 2 in, and price lower than 60.00 dollars
'' and ``Find me home office furniture sets for dining room, living room with color: blue, and item shape: round, and size: 1 ft 9 in x 7 ft 2 in, and price lower than 60.00 dollars''.

\bigskip
Act 1: reset

Obs 1: {`has\_search\_bar': True, `clickables': [`search']}

Act 2: search[home office furniture sets]

Obs 2: {`has\_search\_bar': False, `clickables': [`back to search', `next >', `b09688c4xm', `b09jc84p1m', `b099yq75wt', `b01n4qb5wp', `b09m6vxd6w', `b0048u51n4', `b07tnktkf4', `b09n8slfrj', `b000gltj3m', `b08lnfnxsd']}

Act 3: click[b09jc84p1m]

Obs 3: {`has\_search\_bar': False, `clickables': [`back to search', `< prev', `description', `features', `reviews', `buy now']}

Act 4: click[back to search]

Obs 4: {`has\_search\_bar': True, `clickables': [`search']}

Act 5: search[home office furniture sets navy red round 21 in x 35 in price < 60.00]

Obs 5: {`has\_search\_bar': False, `clickables': [`back to search', `next >', `b07fkgqkz1', `b09jc84p1m', `b07fvhpk5x', `b09688c4xm', `b08p8lrfz4', `b09byx42dw', `b085rbzchh', `b09c1ywg8w', `b09cq45zrb', `b06xfzxxtc']}

Act 6: click[b09cq45zrb]

Obs 6: {`has\_search\_bar': False, `clickables': [`back to search', `< prev', `description', `features', `reviews', `buy now', `15.7x23.6in+15.7x47.2in', `19.7x31.5in+19.7x47.2in', `19.7x31.5in+19.7x63in', `20x24in+20x48in', `23.6x35.4in+23.6x70.9in', `cargoo5209', `christmas-005goo7317', `christmas-010goo9911', `christmasgoo1729', `christmasgoo3302', `christmasgoo3848', `christmasgoo6658', `easter3540lgoo6518', `eastergoo0555', `luckygoo4356', `luckygoo4796', `rabbit3555lgoo1957', `snowmangoo0240', `spt-026goo3844']}

Act 7: click[buy now]

\bigskip
\colorbox{red!15}{Success flag = False, Success rate = 0.25}}
\\ \hline
\end{tabular}
\end{table*}